# Static, dynamic and stability analysis of multi-dimensional functional graded plate with variable thickness using deep neural network

Nam G. Luu[*1] and Thanh T. Banh[2a]

[1] Department of Data Science, Faculty of Information Technology, Industrial University Ho Chi Minh city, 12 Nguyen Van Bao, Go Vap, Ho Chi Minh city 70000, Vietnam
[2] Department of Computer Science, Sejong University, 209 Neungdong-ro, Gwangjin-gu, Seoul 05006, Republic of Korea

**Abstract.** The goal of this paper is to analyze and predict the central deflection, natural frequency, and critical buckling load of the multi-directional functionally graded (FG) plate with variable thickness resting on an elastic Winkler foundation. First, the mathematical models of the static and eigenproblems are formulated in great detail. The FG material properties are assumed to vary smoothly and continuously throughout three directions of the plate according to a Mori-Tanaka micromechanics model distribution of volume fraction of constituents. Then, finite element analysis (FEA) with mixed interpolation of tensorial components of 4-nodes (MITC4) is implemented in order to eliminate theoretically a shear locking phenomenon existing. Next, influences of the variable thickness functions (uniform, non-uniform linear, and non-uniform non-linear), material properties, length-to-thickness ratio, boundary conditions, and elastic parameters on the plate response are investigated and discussed in detail through several numerical examples. Finally, a deep neural network (DNN) technique using batch normalization (BN) is learned to predict the non-dimensional values of multi-directional FG plates. The DNN model also shows that it is a powerful technique capable of handling an extensive database and different vital parameters in engineering applications.

## 1. Introduction

In recent years, functionally graded (FG) materials have made great strides as one of the advanced and intelligent non-homogeneous composites after these kinds of materials were first discovered Koizumi (1997) by Japanese scientists in the mid-1980s. FG materials are usually made from two material phases: ceramic and metal, whose properties vary smoothly and continuously along particular directions. Metal with good ductility and durability can resist mechanical loads, while ceramic with low thermal conductivity is strongly suitable for standing high-temperature environment Nguyen-Xuan *et al.* (2012). Moreover, undesired stresses discontinuity appearing in layers of laminated composites can be eliminated entirely due to a predefined mathematical function. With this outstanding characteristic, FG material has been had the application potential in engineering fields such as aerospace, automobile, electronics, chemistry, and biomedical engineering (Ilschner 1996, Taheri *et al.* 2014, Radhika *et al.* 2018, Arslan *et al.* 2018, Smith *et al.* 2019).

---

∗Corresponding author, E-mail: luugiangnam.math@gmail.com
[a] Ph.D., E-mail: btthanhk11@gmail.com

In the numerical analysis field, plate theory for both uniform and variable thickness plates have been widely developed to predict the structure responses in the past decades, such as classical plate theory (CPT), refined plate theory (RPT) (Nguyen-Xuan *et al.* 2014), first-order shear deformation plate theory (FSDT) (Nguyen-Xuan *et al.* 2012), generalized shear deformation plate theory (GSDT, Thai *et al.* 2014), and high-order shear deformation plate theory (HSDT, Reddy 2000). Furthermore, due to a good and applicable design, variable thickness plates can be used flexibly in industrial applications, such as maritime, civil, and aviation. In the uni-directional FG (UD-FG) plate with z-direction, extensive studies have been introduced in recent years. For instance, Zhang Zhang (2013) presents modeling and HSDT analysis of UD-FG material rectangular plates based on a physical neutral surface. Analytical solutions for bending, vibration, and buckling of FG plates using a couple of stress-based TSDT and power-law distribution are proposed by Kim and Reddy (2013). Then, a comprehensive review of the various methods employed to study the static, dynamic, and stability behavior of FG plates is presented by Swaminathan et al. (2015). Some different types of FG material, such as sigmoid FG (S-FG) micro-scale plates (Jung and Han 2015), are also presented to deal with bending, vibration, and buckling analysis of both standard and sandwich plates in many publications (Jha et al. 2013, Garg et al 2021).

Although UD-FG materials have been applied widely in recent years, in harsh environments such as advanced machines of modern aerospace shuttles or highspeed vehicles, these materials are not perfectly efficient due to their same distributions in all outer surfaces Nemat-Alla (2003). Consequently, FG plates with material properties varying in two or three dimensions are necessary and effective for such engineering applications, even for ones with variable thickness to save their weight. Then, bi-directional FG (BD-FG) and tri-directional FG (TD-FG) materials are mentioned by many researchers in recent years. For instance, an in-plane BD-FG (IBD-FG) material inhomogeneity plate is investigated for buckling, and free vibration problem with a higher-order shear deformation plate isogeometric analysis by Yin *et al.* (2016). A consistent three-dimensional approach is presented by Xiang et al. (2014) based on the scaled boundary finite element method (FEM) to deal with the free vibration and the mechanical buckling of IBD-FG plates. After that, Do et al. (2017) proposed a new third-order shear deformation plate theory to eliminate the shear-locking effect and shear correction factors for bending and buckling problem of BD-FG plates. The free vibration of a BD-FG circular cylindrical shell is analyzed by Ebrahimi and Najafizadeh (2014) with Voigt and Mori–Tanaka ceramic and metal distribution models. Lieu et al. (2018) shows detailed results of bending and free vibration analyses of IBD-FG plates with variable thickness and two material distribution schemes. Besides, TD-FG plates have received considerable attention in studies (Hughes et al. 2005, Adineh and Kadkhodayan 2017). Especially, Do et al. (2020) predicted behaviors based on modified symbiotic organisms search (mSOS) algorithm of the TD-FG plates with deep neural network to reduce computational cost.

On the other hand, the structures on the elastic foundation (EF) are usually used in civil engineering as aircraft runways, building foundation slabs, railway tracks, etc. There are three popular EFs applied in plate theory: Winkler foundation, Pasternak foundation, and Kerr foundation. There are many papers investigated all three foundations for FG plate, such as (Shahsavari *et al.* 2018, Li *et al.* 2021, Amirpour *et al.* 2016, Alinaghizadeh and Shariati 2016). Furthermore, due to the FG material's characteristics that withstand high temperatures, it is also used as an insulating material to discover. Raju and Rao (1988, 1991) present thermal post-buckling of a square plate resting on an elastic foundation with various boundary conditions. Also, some applications of buckling constraint in an elastic medium and topology optimization are considered by Banh et al. (2017), Hoan and Lee (2017), and Hoan et al. (2020). A Winkler-Pasternak foundation is investigated for non-linear thermal free vibration of pre-buckling, and post-buckled FG plates Taczala et al. (2016). Gunda (2013) presents simple closed-form solutions for thermal post-buckling paths of homogeneous, isotropic, square plate configurations resting on elastic Winkler foundation. Hoang *et al.* (2020) presents effects of non-uniform Pasternak elastic foundation on non-linear thermal dynamics of the simply supported plate reinforced by functionally graded (FG) graphene nanoplatelets.

In the recent industrial revolution, data science and artificial intelligence have become more popular and have much application in many fields, including mechanical engineering. Furthermore, a huge computing cost when constructing FEM or solving linear equations (bending problem) and eigenvalue problems (free-vibration and buckling problems), the deep neural network (DNN) is proposed for predicting directly numerical output (such as non-dimensional values) of multi-directional FG plates without any numerical process. The advantage of a DNN is to easily predict complex nonlinear problems by using training data of simple problems with small errors. There are also some papers that use DNN as an optimal method or as a predicting method in numerical analysis, especially plate theory. Some publications used DNN to present mechanical behaviors such as genetic algorithms (Ye *et al.* 2001), optimization (Wang and Xie 2005), fuzzy logic (Shen *et al.* 2013), and so on. Abambres et al. (2013) use an NN-based formula for the buckling load prediction of I-section cellular steel beams under uniformly distributed vertical loads. Do et al. (2018) improves computational cost enhanced by a deep neural network (DNN) and modified symbiotic organisms search (mSOS) algorithm for optimal material distribution of FG plates. Each dataset is randomly created from analysis, and solutions can directly be predicted by using DNN with a robust metaheuristic mSOS algorithm. With the same optimal method, a TDFG plate is considered using a non-uniform rational B-spline (NURBS) basis function for describing material distribution varying through all three directions.

In this article, a finite element analysis with mixed interpolation of tensorial components of the 4-node (MITC4, Bathe and Dvorkin 1985) is applied to Mindlin–Reissner plate theory resting on the elastic Winkler foundation with material distribution varying through three directions. The material properties of TD-FG materials are described by MoriTanaka micro-mechanical scheme. The influences of boundary conditions, lengthto-thickness ratios, types of the plate (uniform or variable thickness plate), and power indexes on the behavior of the bending, free-vibration, and buckling problems of FG plates are considered with various elastic Winkler parameters. After verifying the results with the reference results, all numerical results are collected as training data, and the DNN is considered as an advanced technique to directly predict behaviors (non-dimensional deflection, natural frequency, and critical buckling loads) of the multi-directional FG plates without solving linear equations or eigenvalue problems. Batch normalization accelerates deep network training by requiring lower learning rates and careful parameter initialization. FSDT-based FEM randomly creates these datasets with MITC4 through iterations. The optimal results attained by DNN are compared with those gained by the traditional method to verify the proposed method's effectiveness in both accuracy and time consumption.

The rest of this study is organized as follows. Section 2 exhibits a theoretical formulation of modeling TD-FG plates with the Mori-Tanaka scheme embedded on the elastic Winkler foundation. A construction of FEM with MICT4 for FSDT and the total energy equation are elaborately derived in Sec. 3. Section 4 proposes the deep neural network for predicting the behaviors of TD-FG plates in various variables. Section 5 demonstrates the efficiency and reliability of the present method through several numerical examples of bending, free-vibration, and buckling problems. Finally, Section 6 ends this paper with concluding remarks.

## 2. Multi-directional FG variable thickness plates

In this work, the ceramic volume fraction distribution varying according to a power-law form is first considered as a test in analysis which is represented by the following model.

### 2.1 Numerical simulation procedure

One can write the extended form of the Hamilton's Principle with the notations used in the present study as……

$$V_c(x,y,z) = \left(\frac{x}{a}\right)^{k_x} \left(\frac{y}{a}\right)^{k_x} \left(\frac{1}{2} + \frac{z}{h(x,y)}\right)^{k_z} \tag{1}$$

where a, is size of square plate and $h(x,y) = h_0 \lambda(x,y)$ is variable thickness of the plate; $k_x$, $k_y$ and $k_z$ represent the power indexes in the x-, y- and z-axes, respectively. The function $\lambda(x,y)$ is defined as (Banh and Lee 2019, Banh et al. 2020):

$$\lambda(x,y) = \begin{cases} 1 & \text{type 1: uniform thickness} \\ 1+x & \text{type 2: non-uniform linear thickness} \\ 1+\left(x-\frac{a}{2}\right)^2 + \left(y-\frac{a}{2}\right)^2 & \text{type 3: non-uniform non-linear thickness} \end{cases} \tag{2}$$

Fig. 1 shows the geometrical figures of uniform, non-uniform linear, and nonuniform non-linear thickness plates.

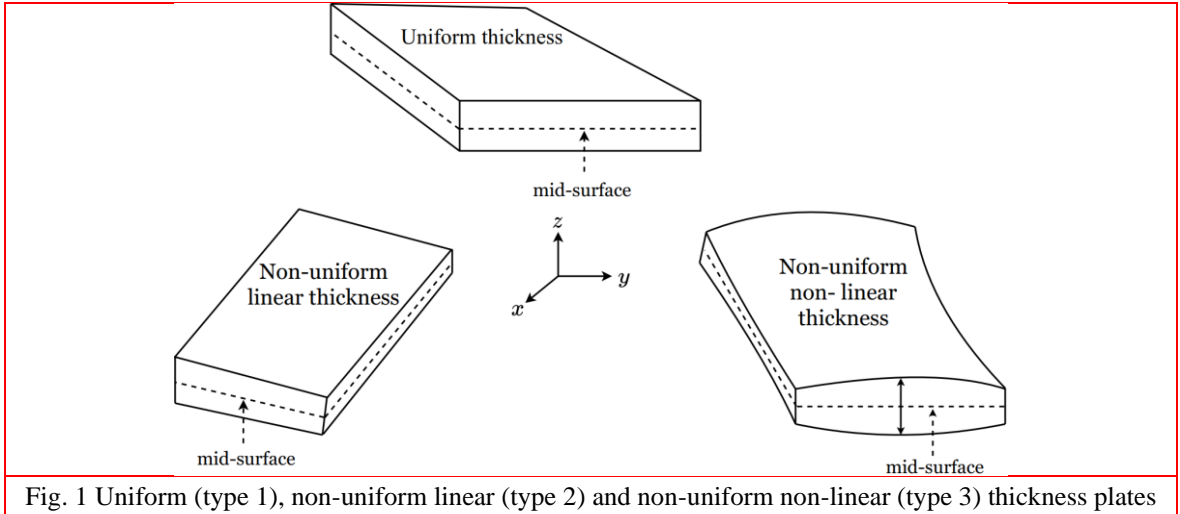
Fig. 1 Uniform (type 1), non-uniform linear (type 2) and non-uniform non-linear (type 3) thickness plates

The metal volume fraction distribution is then computed as follows.

$$V_m(x, y, z) = 1 - V_c(x, y, z) \tag{3}$$

The material properties of the FG plate consisting of Young's modulus E, Poisson's ratio ν, density ρ can be exhibited by the rule of mixture Reddy (2000).

$$P(x, y, z) = P_c \cdot V_c(x, y, z) + P_m \cdot V_m(x, y, z) \tag{4}$$

However, the interactions between two constituents are not taken into account by the rule of mixture (Vel and Batra 2002, Qian *et al.* 2004). As a result, the Mori-Tanaka scheme was used to capture these interactions, wherein the effective bulk and shear modulus can be expressed by.

$$K_f = \frac{V_c(K_c - K_m)}{1 + V_m \frac{K_c - K_m}{K_m + 4/3\mu_m}} + K_m, \mu_f = \frac{V_c(\mu_c - \mu_m)}{1 + V_m \frac{\mu_c - \mu_m}{\mu_m + f_1}} + \mu_m \tag{5}$$

where $f_1 = \frac{\mu_m(9K_m + 8\mu_m)}{6(K_m + 2\mu_m)}$, and $K_{c,m}$ and $\mu_{c,m}$ are bulk and shear moduli of the two phases, respectively, which are defined as

$$K_{c,m} = \frac{E_{c,m}}{3(1 - 2\mu_{c,m})}, \mu_{c,m} = \frac{E_{c,m}}{2(1 + \mu_{c,m})} \tag{6}$$

Then, the effective material properties of the FG plate are now estimated by

$$E = \frac{9K_f \mu_f}{3K_f + \mu_f}, v = \frac{3K_f - 2\mu_f}{2(3K_f + \mu_f)} \tag{7}$$

and ρ(x, y, z) is still computed by Eq. (4).

An FG plate consisting of ceramic and metal phases is depicted in Fig. 2a while the Young modulus distribution of UD-FG SUS304/Si$_3$N$_4$ is presented in Fig. 2b.

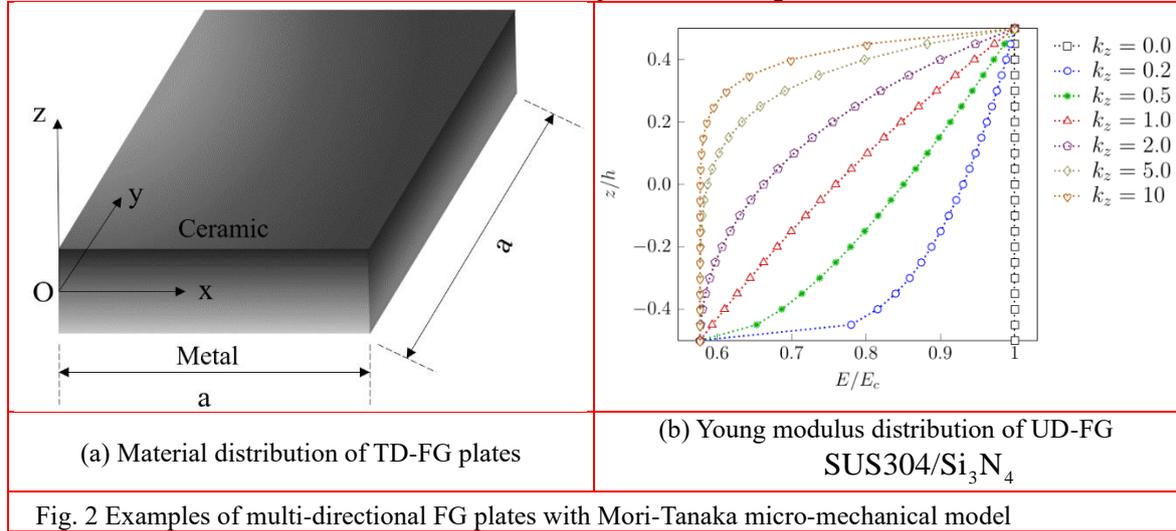

(a) Material distribution of TD-FG plates

(b) Young modulus distribution of UD-FG SUS304/Si$_3$N$_4$

Fig. 2 Examples of multi-directional FG plates with Mori-Tanaka micro-mechanical model

## 3. Basic formulations for FSDT using FEA and MITC4

*3.1 Formulation of FSDT finite element model*

According to FSDT, the in-plane displacement field at an arbitrary point (x, y, z) of a plate can be approximated as

$$u(x, y, z) = u_0(x, y, t) + z\phi_x(x, y)$$
$$v(x, y, z) = v_0(x, y, t) + z\phi_y(x, y) \qquad (8)$$
$$w(x, y) = w_0(x, y)$$

where $u_0$, $v_0$, $w_0$, $\phi_x$ and $\phi_y$ are function of x, y, t (time); $u_0$, $v_0$, $w_0$ denote the displacements of a point on the middle surface; $\phi_x$ and $\phi_y$ are the rotations of a transverse normal about the y-, x-axis. Note that, a comma represents the differentiation to the space. Fig. 3 shows the deformed plate cross-section view.

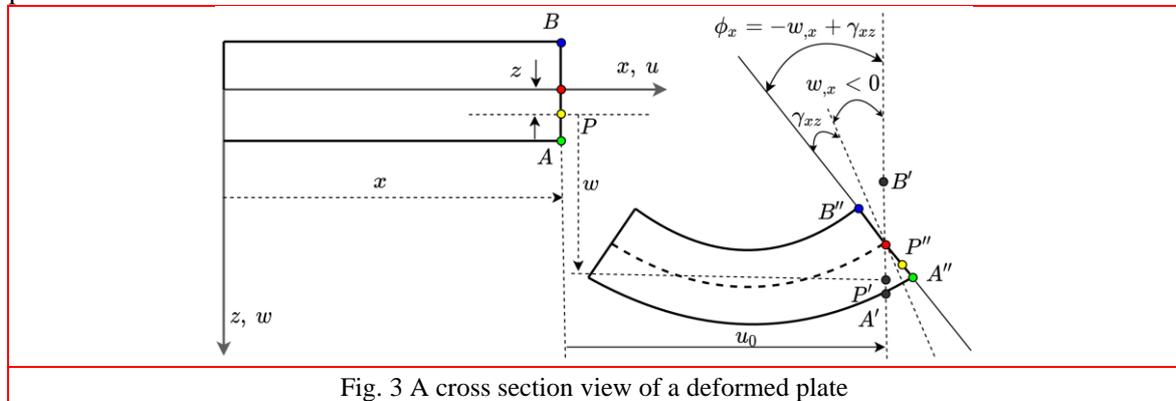

Fig. 3 A cross section view of a deformed plate

The generalized displacements and rotations of the plates using the FEA can be approximated: with the bilinear quadrilateral shape functions $N_i$. Then, strain is expressed based on FEA as follows:

$$\mathbf{u} = \sum_{i=1,4} N_i \mathbf{q}_i, \text{where } \mathbf{u} = \{u, v, \beta_x, \beta_y\}^T, \mathbf{q}_i = \{u_i, v_i, \beta_{xi}, \beta_{yi}\}^T \tag{8}$$

$$\boldsymbol{\varepsilon}_0 = \{u_{0,x} \quad v_{0,y} \quad u_{0,y} + v_{0,x}\}^T, \qquad \boldsymbol{\varepsilon}_1 = \{\phi_{x,x} \quad \phi_{y,y} \quad \phi_{x,y} + \phi_{y,x}\}^T$$

$$\{\boldsymbol{\varepsilon}_0 \quad \boldsymbol{\varepsilon}_1\}^T = \sum_{i=1,4} \mathbf{B}_i^{mb} = \mathbf{B}^{mb}\mathbf{q}, \qquad \boldsymbol{\varepsilon}_2 = \{\phi_x + w_{0,x} \quad \phi_y + w_{0,y}\}^T = \mathbf{B}^s \mathbf{q}$$

$$\mathbf{B}^k = \{\mathbf{B}_1^k \quad \mathbf{B}_2^k \quad \mathbf{B}_3^k \quad \mathbf{B}_4^k\}^T, k \in \{mb,s\} \qquad \mathbf{q} = \{\mathbf{q}_1 \quad \mathbf{q}_2 \quad \mathbf{q}_3 \quad \mathbf{q}_4\}^T \tag{9}$$

$$\mathbf{B}_i^{mb} = \begin{bmatrix} N_{i,x} & 0 & 0 & 0 & 0 \\ 0 & N_{i,y} & 0 & 0 & 0 \\ N_{i,y} & N_{i,x} & 0 & 0 & 0 \\ 0 & 0 & 0 & N_{i,x} & 0 \\ 0 & 0 & 0 & 0 & N_{i,y} \\ 0 & 0 & 0 & N_{i,y} & N_{i,x} \end{bmatrix}, \quad \mathbf{B}_i^s = \begin{bmatrix} 0 & 0 & N_{i,x} & N_{i,y} & 0 \\ 0 & 0 & N_{i,y} & 0 & N_{i,y} \\ 0 & 0 & 0 & N_{i,y} & N_{i,x} \end{bmatrix}$$

Based on the generalized Hoole's law, the stress is determined by the constitutive relations as

$$\{\sigma_{xx} \quad \sigma_{yy} \quad \sigma_{xy}\}^T = \mathbf{Q}^{mb}\{\varepsilon_{xx} \quad \varepsilon_{yy} \quad \varepsilon_{xy}\}^T, \{\tau_{xz} \quad \tau_{yz}\}^T = \mathbf{Q}^s\{\gamma_{xz} \quad \gamma_{yz}\}^T \tag{10}$$

where $\{\varepsilon_{xx} \quad \varepsilon_{yy} \quad \varepsilon_{xy}\}^T = \boldsymbol{\varepsilon}_0 + z\boldsymbol{\varepsilon}_1, \{\gamma_{xz} \quad \gamma_{yz}\}^T = \boldsymbol{\varepsilon}_2$ and

$$\mathbf{Q}^{mb} = \begin{Bmatrix} Q_{11} & Q_{12} & 0 \\ Q_{21} & Q_{22} & 0 \\ 0 & 0 & Q_{44} \end{Bmatrix}, \mathbf{Q}^s = \begin{Bmatrix} Q_{55} & 0 \\ 0 & Q_{66} \end{Bmatrix} \tag{11}$$

$$Q_{11} = Q_{22} = \frac{E}{1-v^2}, Q_{12} = Q_{21} = \frac{Ev}{1-v^2}, Q_{44} = Q_{55} = Q_{66} = \frac{E}{2(1+v)}$$

The stress resultants are related to the strains by the following relationships

$$\{N_{\alpha,\beta}, M_{\alpha,\beta}\} = \int_{-h/2}^{h/2} \{1, z\} \sigma_{\alpha\beta} dz, S_\alpha = \int_{-h/2}^{h/2} \tau_{\alpha z} dz \tag{12}$$

where $\{\alpha, \beta\} = \{x, y\}$. From Eqs. (10) and (12) it can be computed that

$$\left\{\begin{matrix} \mathbf{N} \\ \mathbf{M} \\ \mathbf{S} \end{matrix}\right\} = \begin{bmatrix} \mathbf{D}^{mb} & [\mathbf{0}] \\ [\mathbf{0}] & \mathbf{D}^{s} \end{bmatrix} \left\{\begin{matrix} \boldsymbol{\varepsilon}_0 \\ \boldsymbol{\varepsilon}_1 \\ \boldsymbol{\varepsilon}_2 \end{matrix}\right\} = \mathbf{D}^{mb} \left\{\begin{matrix} \boldsymbol{\varepsilon}_0 \\ \boldsymbol{\varepsilon}_1 \end{matrix}\right\} + \mathbf{D}^{s} \boldsymbol{\varepsilon}_2 = \mathbf{D}^{mb} \mathbf{B}^{mb} \mathbf{q} + \mathbf{D}^{s} \mathbf{B}^{s} \mathbf{q} \tag{13}$$

where

$$\mathbf{D}^{mb} = \begin{bmatrix} \mathbf{A}^1 & \mathbf{A}^2 \\ \mathbf{A}^2 & \mathbf{A}^3 \end{bmatrix}, \mathbf{D}^{s} = \int_{-h/2}^{h/2} \mathbf{Q}^{s} dz \tag{14}$$

$$A_{ij}^1 = \int_{-h/2}^{h/2} \mathbf{Q}^{mb} dz, A_{ij}^2 = \int_{-h/2}^{h/2} z \mathbf{Q}^{mb} dz, A_{ij}^3 = \int_{-h/2}^{h/2} z^3 \mathbf{Q}^{mb} dz \tag{15}$$

### *3.2 Formulation of MITC4 scheme*

The construction of thick plate elements, in which the deflection and the rotation are independently defined, is indeed more straightforward than that of thin plate elements to develop bending components based on the Mindlin-Reissner plate theory (Zhang and Kuang 2007). However, in the thin plate case from the low-order standard iso-parametric displacement-based plate elements without unique treatments only, the instability in shear strains of model Kirchhoff-type constraints in thin plate limits is occurred due to the shear locking phenomenon (Arnold 1981). Therefore, shear energy is defined in terms of the assumed covariant transverse shear strain field of the MITC4 (Thompson and Thangavelu 2002) to eliminate the shear locking phenomenon in the FSDT plate theory based on the approximated membrane-bending part. However, the approximation of the shear strain part has to be re-defined by the linear interpolation between mid-points of the element edges, namely that assume the transversal shear interpolation in local convective coordinates to be linear in ν direction as shown in Fig. 4.

$$\boldsymbol{\gamma}_{\xi\nu} = \begin{bmatrix} \gamma_\xi \\ \gamma_\nu \end{bmatrix} = \frac{1}{2} \begin{bmatrix} (1-\nu)\gamma_\xi^C + (1+\nu)\gamma_\xi^A \\ (1-\nu)\gamma_\xi^B + (1+\nu)\gamma_\xi^D \end{bmatrix} \tag{16}$$

$$\boldsymbol{\gamma}_{xy} = \begin{bmatrix} \gamma_x \\ \gamma_y \end{bmatrix} = \begin{bmatrix} \xi_{,x} & \nu_{,x} \\ \xi_{,y} & \nu_{,y} \end{bmatrix}^{-1} \begin{bmatrix} \gamma_\xi \\ \gamma_\nu \end{bmatrix} = \mathbf{J}^{-1} \begin{bmatrix} \gamma_\xi \\ \gamma_\nu \end{bmatrix} \tag{17}$$

where the strain components at points A, B, C and D, to obtain:

$$\gamma_\psi^\chi = x_{,\psi}^\chi \frac{\phi_x^a + \phi_x^b}{2}, + y_{,\psi}^\chi \frac{\phi_y^a + \phi_y^b}{2} + \frac{w_a - w_b}{2} \tag{18}$$

in which $(\xi,\psi,a,b) = \{(\xi,C,2,1),(\xi,A,3,4),(\nu,B,4,1),(\nu,D,3,2)\}$.

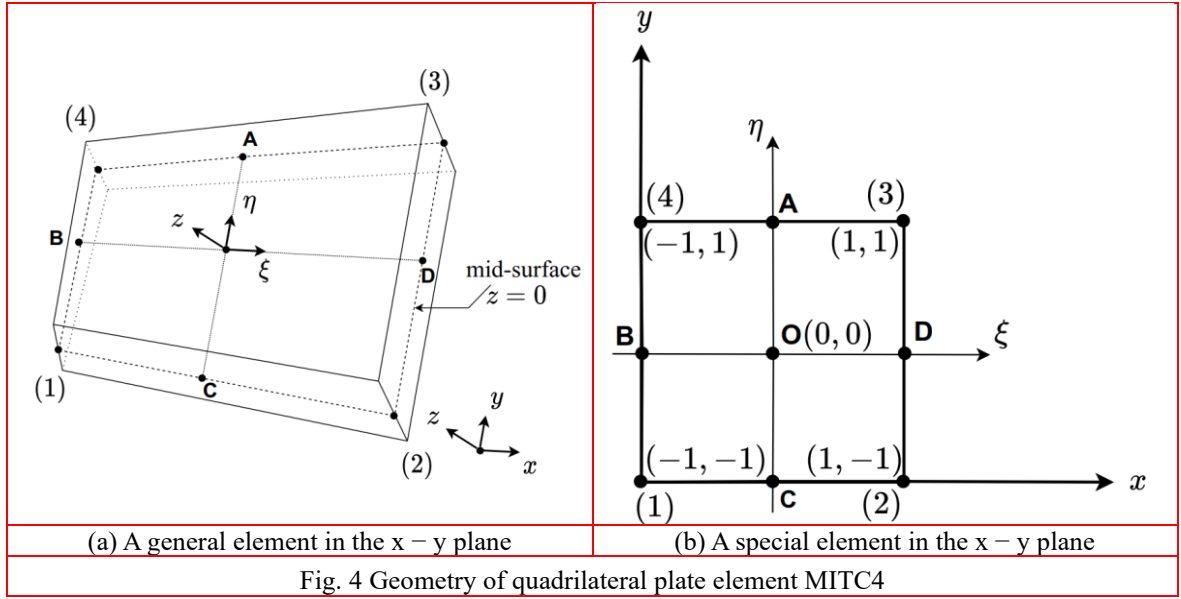

| (a) A general element in the x − y plane | (b) A special element in the x − y plane |

Fig. 4 Geometry of quadrilateral plate element MITC4

By using the MITC4 technique, the approximation of modified shear strain components may be expressed as

$$\gamma^m = \mathbf{J}^{-1} \left\{ \gamma^1_{\xi\nu} \quad \gamma^2_{\xi\nu} \quad \gamma^3_{\xi\nu} \quad \gamma^4_{\xi\nu} \right\}^T \left\{ w^{(i)} \quad \phi^i_x \quad \phi^i_y \right\} = \mathbf{B}^m \mathbf{q} \tag{19}$$

where

$$\gamma^i_{\xi\nu} = \begin{bmatrix} N^{(i)}_{,\xi} & \xi^{(i)} x^{B,D}_{,\xi} N^{(i)}_{,\xi} & \xi^{(i)} y^{B,D}_{,\xi} N^{(i)}_{,\xi} \\ N^{(i)}_{,\nu} & \eta^{(i)} x^{B,D}_{,\eta} N^{(i)}_{,\eta} & \eta^{(i)} y^{B,D}_{,\eta} N^{(i)}_{,\eta} \end{bmatrix} \tag{20}$$

in which $(i,\chi,\psi) \in \{(1,C,B),(2,C,D),(3,A,D),(4,A,B)\}$ and $\mathbf{J}$ is the Jacobian transformation matrix of mapping $\mathbf{x}:[-1,1]^2 \to \Omega^e$.

### 3.3 Energy and frequency equation

The total strain energy of the initially stressed plate is given by the strain energy due to vibratory stresses as follows.

$$U = \frac{1}{2} \int_{V_e} \left( \sigma_{xx}\varepsilon_{xx} + \sigma_{yy}\varepsilon_{yy} + \sigma_{xy}\varepsilon_{xy} + \tau_{xz}\gamma_{xz} + \tau_{yz}\gamma_{yz} \right) dV_e \tag{21}$$

Applying the finite element model:

$$\delta U = \int_{\Omega_e} \left[ \delta \mathbf{q}^T \left( \mathbf{B}^{mb} \right)^T \mathbf{D}^{mb} \mathbf{B}^{mb} \mathbf{q} + \delta \mathbf{q}^T \left( \mathbf{B}^s \right)^T \mathbf{D}^s \mathbf{B}^s \mathbf{q} \right] d\Omega_e \tag{22}$$

where $\mathbf{N}^{Th} = \begin{bmatrix} \mathbf{N}^{Th}_1 & \mathbf{N}^{Th}_2 & \mathbf{N}^{Th}_3 & \mathbf{N}^{Th}_4 \end{bmatrix}$ in which

$$\mathbf{N}_i^{\text{Th}} = \begin{bmatrix} N_{i,x} & 0 & 0 & zN_{i,x} & 0 \\ 0 & N_{i,x} & 0 & 0 & 0 \\ 0 & 0 & N_{i,x} & 0 & 0 \\ N_{i,y} & 0 & 0 & zN_{i,y} & 0 \\ 0 & N_{i,y} & 0 & 0 & zN_{i,y} \\ 0 & 0 & N_{i,y} & 0 & 0 \end{bmatrix} \quad (23)$$

The virtual work done by the applied transverse load $q_0$, in-plane buckling forces, the foundation reaction for a typical finite element, and the virtual kinetic energy $\Omega_e$ are denoted $\delta V$, $\delta V_B$, $\delta V_S$ and $\delta K$ given in detail below.

$$\delta V = -\int_{\Omega_e} q_0 \delta w_0 d\Omega_e = -\int_{\Omega_e} \delta \mathbf{q}^T q_0 \mathbf{N}_w d\Omega_e \quad (24)$$

$$\delta V_B = -\int_{\Omega_e} (\nabla \delta w_0)^T \mathbf{N}^0 \nabla w_0 d\Omega_e = -\int_{\Omega_e} (\mathbf{B}_g)^T \mathbf{N}^0 \mathbf{B}_g d\Omega_e \quad (25)$$

$$\delta W_s = k_W \int_{\Omega_e} w_0 \delta w_0 d\Omega_e = \int_{\Omega_e} \delta \mathbf{q}^T (\mathbf{N}_w^T k_W \mathbf{N}_w) \mathbf{q} d\Omega_e \quad (26)$$

$$\delta K = \int_{V_e} \rho(\dot{u}\delta\dot{u} + \dot{v}\delta\dot{v} + \dot{w}\delta\dot{w}) dV_e = \omega^2 \int_{\Omega_e} \delta \mathbf{q}^T \left[ \mathbf{N}_p^T \mathbf{m} \mathbf{N}_p \right] \mathbf{q} d\Omega_e \quad (27)$$

where $k_W$ is the Winkler modulus of subgrade reaction as shown in Fig. 5, and

$$\mathbf{N}_w = \{\mathbf{N}_w^1 \quad \mathbf{N}_w^2 \quad \mathbf{N}_w^3 \quad \mathbf{N}_w^4\}^T, \mathbf{N}_w^i = [0 \quad 0 \quad N_i \quad 0 \quad 0] \quad (28a)$$

$$\mathbf{B}_g = \{\mathbf{B}_g^1 \quad \mathbf{B}_g^2 \quad \mathbf{B}_g^3 \quad \mathbf{B}_g^4\}^T, \mathbf{B}_g^i = \begin{bmatrix} 0 & 0 & N_{i,x} & 0 & 0 \\ 0 & 0 & N_{i,y} & 0 & 0 \end{bmatrix} \quad (28b)$$

$$\mathbf{N}_p = \{\mathbf{N}_p^1 \quad \mathbf{N}_p^2 \quad \mathbf{N}_p^3 \quad \mathbf{N}_p^4\}^T, \mathbf{N}_p^i = \{\mathbf{N}_p^{i1} \quad \mathbf{N}_p^{i2} \quad \mathbf{N}_p^{i3}\}^T \quad (28c)$$

$$\mathbf{N}_p^{i1} = \begin{bmatrix} N_i & 0 & 0 & 0 & 0 \\ 0 & 0 & 0 & N_i & 0 \end{bmatrix}, \mathbf{N}_p^{i2} = \begin{bmatrix} 0 & N_i & 0 & 0 & 0 \\ 0 & 0 & 0 & 0 & N_i \end{bmatrix} \quad (28d)$$

$$\mathbf{N}_p^{i3} = \begin{bmatrix} 0 & 0 & N_i & 0 & 0 \\ 0 & 0 & 0 & N_i & 0 \end{bmatrix}, \mathbf{N}_0 = \begin{bmatrix} N_{xx}^0 & N_{xy}^0 \\ N_{xy}^0 & N_{yy}^0 \end{bmatrix} \quad (28e)$$

$$\mathbf{m} = \begin{bmatrix} \mathbf{m}_0 & \mathbf{0} & \mathbf{0} \\ \mathbf{0} & \mathbf{m}_0 & \mathbf{0} \\ \mathbf{0} & \mathbf{0} & \mathbf{m}_0 \end{bmatrix}, \mathbf{m}_0 = \begin{bmatrix} I_0 & I_1 \\ I_1 & I_2 \end{bmatrix} \tag{28f}$$

$$I_0 = \int_{-h/2}^{h/2} \rho(x,y,z)dz, I_1 = \int_{-h/2}^{h/2} z\rho(x,y,z)dz, I_2 = \int_{-h/2}^{h/2} z^2\rho(x,y,z)dz \tag{28g}$$

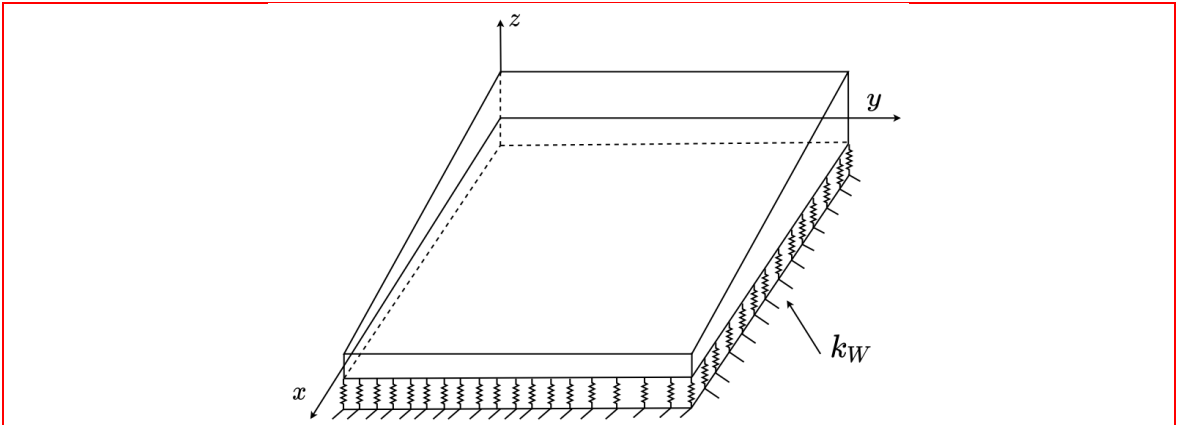

Fig. 5 A non-uniform thickness plate model embedded in elastic foundation presented by a Winkler model

Note that the superposed dot on a variable denotes the time derivative and the above equation is derived based on an assumption of the plate harmonic motion, i.e. $\ddot{\mathbf{u}}_p = -\omega^2 \mathbf{u}_p$ where $\omega$ symbolizes the frequency of natural vibration; $\rho(x, y, z)$ is the mass density/volume; $\mathbf{N}^0$ is a matrix which includes the in-plane applied forces.

Consequently, the finite element model of static bending, free vibration, and buckling problems of TD-FG plate resting on Winkler foundation is respectively expressed in the form of the following linear algebraic equations

$$\mathbf{Kq} = \mathbf{F}, (\mathbf{K} - \omega^2 \mathbf{M})\mathbf{q} = \mathbf{0}, (\mathbf{K} - \lambda_{cr}\mathbf{K}_g)\mathbf{q} = \mathbf{0} \tag{29}$$

where $\mathbf{q}$ is the DOFs vector at the FE analysis, $\mathbf{K}, \mathbf{M}, \mathbf{K}_g$, and $\mathbf{F}$ are the stiffness matrix, element mass matrix, geometric stiffness matrix, and the element applied load vector defined as

$$\begin{aligned} \mathbf{K} &= \int_\Omega \left(\mathbf{B}^{mb}\right)^T \mathbf{D}^{mb}\mathbf{B}^{mb} + \left(\mathbf{B}^s\right)^T \mathbf{D}^s\mathbf{B}^s + \mathbf{N}_w^T k_W \mathbf{N}_w d\Omega_e \\ \mathbf{M} &= \int_{\Omega_e} \mathbf{N}_p^T \mathbf{m} \mathbf{N}_p d\Omega_e, \mathbf{K}_g = \int_{\Omega_e} \mathbf{B}_g^T \mathbf{N}^0 \mathbf{B}_g d\Omega_e, \mathbf{F} = \int_{\Omega_e} \mathbf{N}_w^T q_0 d\Omega_e \end{aligned} \tag{30}$$

## 4. Creating prediction of deep neural network (DNN)

### 4.1 Creating a DNN model

To predict the non-dimensional fundamental frequencies of the plate, we first have to train the ANN model by several sets of material parameters, stiffness of elastic foundation, and non-dimensional fundamental frequencies. The ANN model will be set up to have some input data: volume fraction indexes ($k_x$, $k_y$, $k_z$), stiffness of the foundation $k_W$, length-thickness ratio ($a/h_0$), type of problem; the output is the non-dimensional value based on type of problem: central deflection $\bar{w}$) for bending, fundamental frequencies $\bar{\omega}$) for free-vibration and critical buckling load ($\bar{\lambda}_{cr}$) for uni- and bi-axial buckling problem.

### 4.2 Collecting data for training

After completing to implement FEM with MITC4 to deal with TD-FG variable thickness plates, the present approach is verified with high accuracy with reference papers. Then, all results will be collected as a sample of DNN model with input variables: 'problem'(bending, free-vibration, uni-, and bi-buckling), 'boundary condition' (CCCC, SSSS, CFCF, SFSF, CSCS where 'C', 'S', and 'F' stand for clamped, simply supported, and free conditions, respectively), $k_x$, $k_y$, $k_z$ and $k_W$, type of plate (as shown in Fig. 1), $a/h_0$ and the results are their non-dimensional data. A TD-FG SUS304/$Si_3N_4$ is used in all training and test problems. For the training process in the deep neural network, dataset training patterns are randomly created through iterations from a general code of TD-FG variable thickness plate under elastic foundation. There are 20,000 training patterns designed for the training set for the TD-FG problem with random inputs. Note that 90% data patterns are randomly chosen to train data, and 10% left is used for the validation process. By using the dataset, DNN performs the training process through 5,000 epochs to obtain the optimal mapping rules expressed by weights.

### 4.3 Theoretical DNN model

A multi-layer perceptron (MLP) model architecture is utilized to predict nondimensional values. In this article, the model consists of three hidden layers with the number of neurons in the inputs mentioned earlier. As the model is quite shallow (only three hidden layers), a Sigmoid function is selected as an activation function instead of ReLU or its variants despite their popularity. Adam optimizer (Kingma and Ba 2014) is applied as an adaptive learning rate optimization algorithm that's been designed specifically for training DNN. Loss function is defined as:

$$L = MSE = \frac{1}{n}\sum_{i=1,n}\left(Y_i - Y_i\right)^2 \quad (31)$$

where MSE is mean square error, n is batch size, $Y_i$ is ground truth, and $Y_i$ is prediction.

In order to make the training of NN faster and more stable, an algorithmic method named Batch-Normalization (BN, Ioffe and Szegedy 2015) is applied. BN consists of normalizing activation vectors from hidden layers using the current batch's first and second statistical moments (mean and variance). This normalization step is applied right before (or right after) the non-linear function. BN is computed differently during the training and the testing phase. At each hidden layer, BN transforms the signal as follow:

$$Z = \frac{Z^{(i)} - \mu_i}{\sqrt{\sigma_i^2 + \varepsilon}} * \gamma_i + \beta_i, \mu = \frac{1}{n}\sum_i Z^{(i)}, \sigma^2 = \frac{1}{n}\sum_i \left(Z^{(i)} - \mu\right)^2 \quad (31)$$
$$Z_{norm}^{(i)} = \left(Z^{(i)} - \mu\right)/\sqrt{\sigma^2 + \varepsilon}$$

The BN layer first determines the mean μ and the variance $\sigma^2$ of the activation values across the batch. It then normalizes the activation vector $Z^{(i)}$. That way, each neuron's output follows a standard normal distribution across the batch. It finally calculates the layer's output $\hat{Z}$ by applying a linear transformation with γ and β, two trainable parameters $\hat{Z} = \gamma Z_{norm}^{(i)} + \beta$. In each hidden layers, the optimum distribution is chosen for the DNN model. At each iteration, the network computes the mean μ and the standard deviation σ corresponding to the current batch and trains γ and β. The input data are pre-processed before being fed to the neural network model. Non-scalar value is one-hot encoded, while scalar value is normalized with Z-Score Normalization: new value = $\left(Z^{(0)} - \mu_0\right)/\sigma_0$, where $Z^{(0)}$ is original value, μ is mean of data, and σ is standard deviation of data.

Table 1 shows the advantages of BN in the Multilayer Perceptron (MLP). Table 2 shows comparison of Adam optimizers with different activation functions (Sigmoid, Softplus, ReLU, and Tanh). As can be seen, the best option for both training error and validation error is Sigmoid function with BN.

Table 1 Comparisons of Multilayer Perceptron with and without batch normalization.

| Without batch normalization | With batch normalization |
|---|---|
| • Raw signal | • Normalized signal |
| • High interdependency between distribution | • Mitigated interdependency between distribution |
| • Slow and unstable training | • Fast and stable training |

Table 2 Mean square error for training and valid sets of Adam optimizer with activation functions.

| | Without batch normalization | | | | With batch normalization | | | |
|---|---|---|---|---|---|---|---|---|
| | Sigmoid | Softplus | ReLU | Tanh | Sigmoid | Softplus | ReLU | Tanh |
| Train | 1.84% | 1.41% | 0.52% | 0.50% | 0.72% | 0.72% | 0.75% | 0.59% |
| Valid | 4.54% | 2.71% | 1.92% | 1.36% | 1.19% | 1.31% | 3.54% | 1.83% |

The DNN model is illustrated in Fig. 6 while Fig. 7 shows the convergence history of training loss function and valid loss function achieved by the above-mentioned DNN model. A flowchart of the DNN approach for predicting non-dimensional values of variable thickness TD-FG plate problems resting on an elastic Winkler foundation is depicted in Fig. 8.

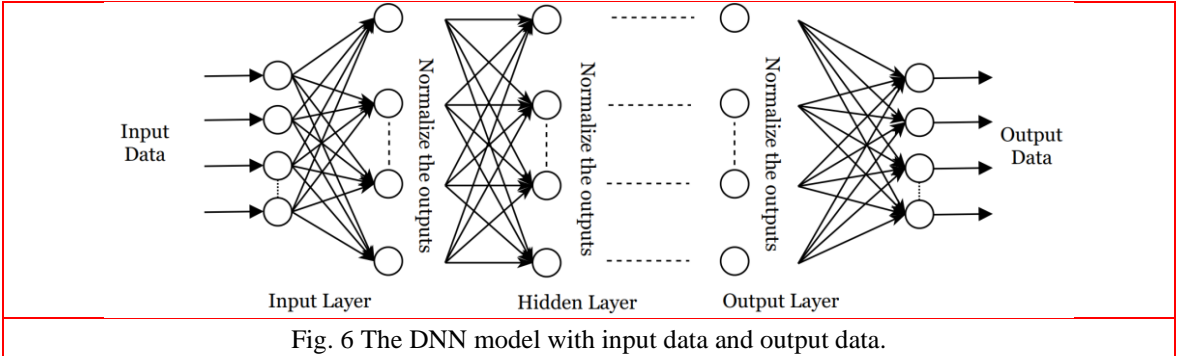
Fig. 6 The DNN model with input data and output data.

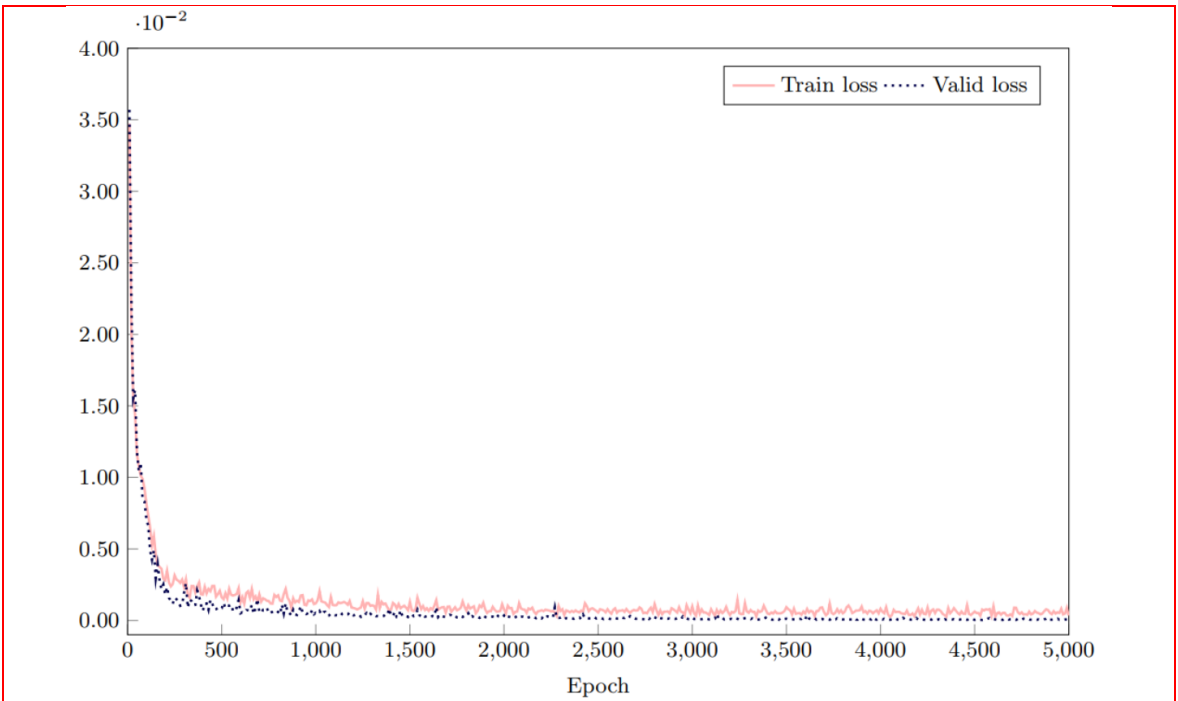
Fig. 7 The convergence history of train and valid loss functions with random dataset

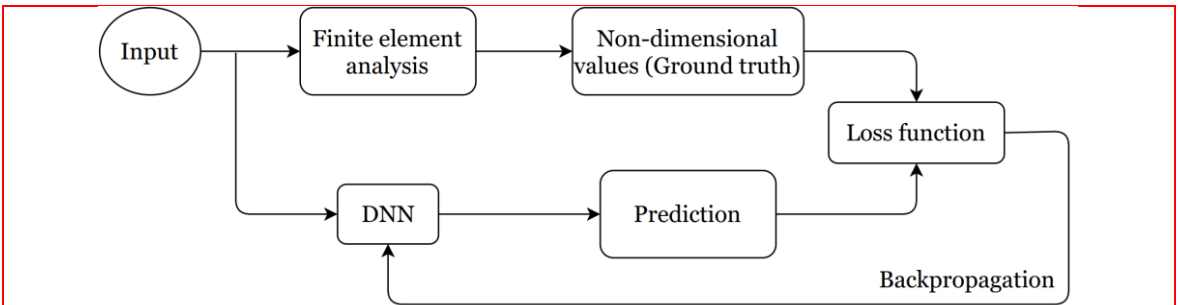
Fig. 8 A computational flowchart of DNN for predicting behavior of static, dynamic, and stability problems

## 5. Numerical examples

In this study, bending, free vibration and buckling behaviours of some types of multi-directional FG plates are investigated to verify the proposed method with the previous papers and present the new results with DNN prediction. At first, material properties of these FG plates are presented in Tab. 3:

Table 3 Information of normal FG material.

|  | Material | E(Pa) | $v$ | $\rho\left(kg/m^3\right)$ |
|---|---|---|---|---|
| Ceramics | Silicon nitride (Si$_3$N$_4$) | 348.43 | 0.24 | 2370 |
|  | Alumina (Al$_2$O$_3$) | 380 | 0.3 | 3800 |
|  | Zirconia (ZrO$_2$) | 151 | 0.3 | 3000 |
| Metals | Stainless steel (SUS304) | 201.04 | 0.3262 | 8166 |
|  | Aluminum (Al) | 70 | 0.32 | 2702 |

The properties of FG materials are estimated the Mori-Tanaka scheme and all uni-, bi-, and tri-directional FG are surveyed in each linear algebra and eigenvalue problem. Moreover, the effects of gradient indexes, boundary conditions, length-to-thickness ratios, temperature in ceramic phase (only thermal case), and elastic Winkler parameter on multi-directional FG plates are also considered. In all problems in this section, the shear correction factor is fixed at SCF $=\pi^2/12$. After comparing to the reference result, all non-dimensional values according to bending problem (central deflection $\bar{w}$), free-vibration problem (natural frequency $\bar{\omega}$) and uni- and bi-axial buckling problem (critical buckling load $\bar{P}_{cr}$) are calculated for present study and collecting data with FG material SUS304/Si$_3$N$_4$ as follow:

$$\bar{w}\left(\frac{a}{2},\frac{a}{2},0\right) = w\left(\frac{a}{2},\frac{a}{2},0\right)\frac{E_c h_0^2}{a^3 q_0} \tag{33a}$$

$$\bar{\omega} = \omega\left(a/\pi\right)^2 \sqrt{\frac{\rho_c h_0}{D_c}} \tag{33b}$$

$$\bar{P}_{cr} = \frac{P_{cr} a^2}{\pi^2 D_c} \tag{33c}$$

where $D_c = E_c h_0^3 / 12\left(1-v_c^2\right)$.

### 4.1 Bending problems

**Example 1:** In this example, uniform thickness UD-FG (Al/ZrO$_2$) thick plates (k$_x$=k$_y$=0) with a length-to-thickness ratio a/h$_0$ = 5 are investigated using of three different boundary conditions CCCC, SSSS, and SFSF. The load distribution in this example is considered as a uniform load with the magnitude value q=q$_0$=-1 (see Fig. 9a). The isotropic material properties are assumed to according to the Mori-Tanaka model distribution of volume fraction of constituents. The non-dimensional transverse displacement can be defined as

$$\bar{w}\left(\frac{a}{2},\frac{a}{2},0\right)=100w\left(\frac{a}{2},\frac{a}{2},0\right)\frac{q_0 E_m h_0^3}{12a^4\left(1-v_m^2\right)} \quad (34)$$

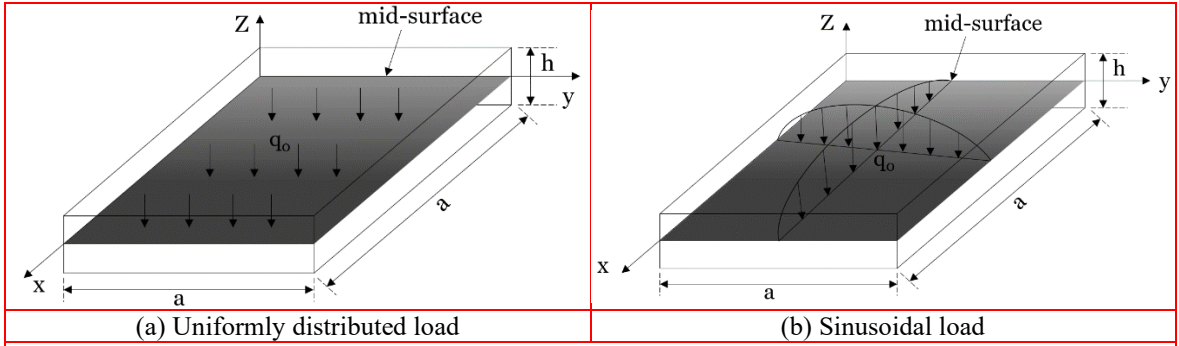

(a) Uniformly distributed load      (b) Sinusoidal load

Fig. 9 Geometry of FG plate under uniformly distributed and sinusoidal loads

Fig. 10 shows the non-dimensional transverse displacement compared with those of GSDT (Nguyen-Xuan *et al.* 2014) and TSDT (Tran *et al.* 2013) for UD-FG plate of bending problem with SSSS and CCCC boundary conditions. From the figure, it can be seen that the gained transverse displacement is in good agreement with the existing results with a small percentage relative error.

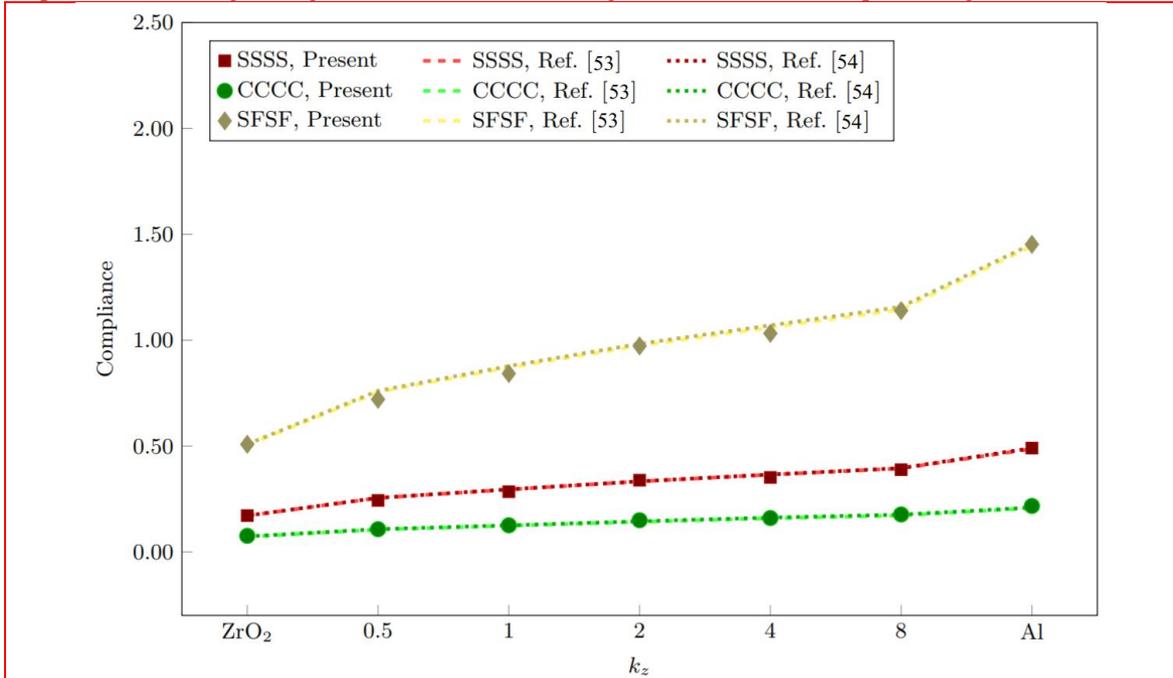

Fig. 10 The comparison of non-dimensional transverse displacement of UD-FG Al/$ZnO_2$ plate under uniform load in bending problem

**Example 2:** An in-plane BD-FG SUS304/$Si_3N_4$ ($k_z$=0) square plate subjected to a sinusoidal load of $q=q_0\sin(\pi x/a)\sin(\pi y/a), q_0=1$ (as depicted in Fig. 9b) is considered in this example. The non-dimensional central deflection ($\bar{w}$ in Eq. 33a) of the plate is examined with CCCC and SSSS boundary conditions and compared with Lieu et al. (2018). Fig. 11 illustrates the comparison of the above problem with previous in case of thick plate ($a/h_0=5$) and thin plate ($a/h_0=100$). From the figure, it can be observed that obtained results by the present method are in excellent agreement with the reference results.

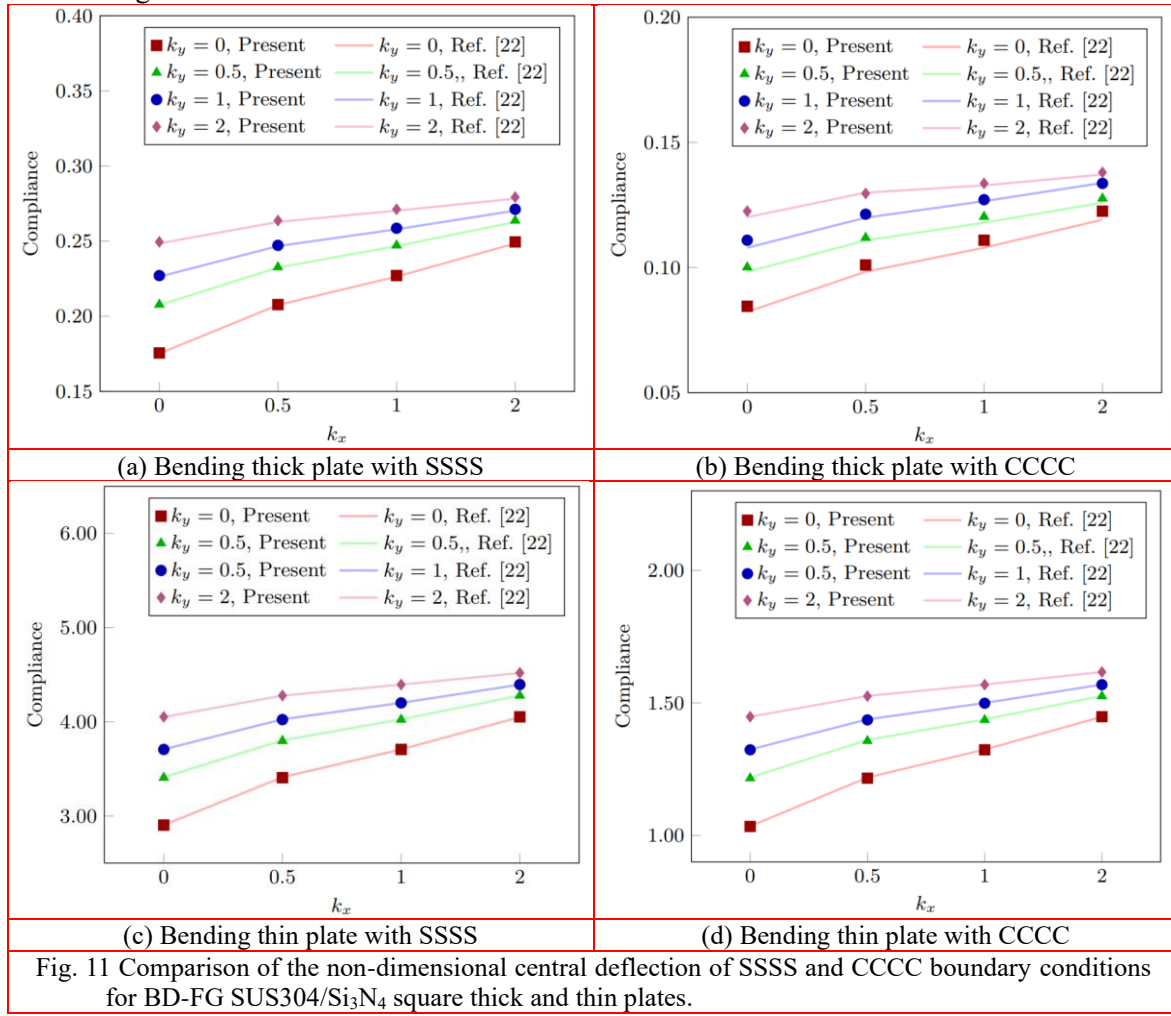

Fig. 11 Comparison of the non-dimensional central deflection of SSSS and CCCC boundary conditions for BD-FG SUS304/Si$_3$N$_4$ square thick and thin plates.

From above examples, it can be concluded that the present analytical approach for the mechanical problem is effective and reliable for analysing behaviour of the FG plate in term of bending problem. Thus, the present method may ensure the generated dataset for the training process in the DNN regarding the bending problem in the subsequent example.

**Example 3:** A TD-FG SUS304/Si$_3$N$_4$ square plate subjected to sinusoidal loads, as depicted in Fig. 9b, is considered in this example. The non-dimensional central deflection Eq. (33a) of the plate is examined with CCCC and SSSS boundary conditions. Table 4 demonstrates the present study of the above problem with DNN prediction in case of thick plate ($a/h_0=20$) and thin plate ($a/h_0=50$). It can be seen that the DNN prediction is closed to the present study using finite element analysis with MITC4 that verifies the accuracy of DNN model. The maximum relative error in TD-FG bending problem is 2.5492% in the thick plate $a/h_0=20$ with CCCC $k_x=8$, $k_y=4$, $k_z=2$ (FEM result is 0.3550, DNN prediction is 0.3529). The average relative error in this case is 1.3065%. Fig. 12 shows the deflection of TD-FG thick and thin plates with CCCC and SSSS and $k_x=4$, $k_y=4$, $k_z=2$.

Table 4 The non-dimensional central deflection $\bar{w}\left(\frac{a}{2},\frac{a}{2},0\right) = w\left(\frac{a}{2},\frac{a}{2},0\right)\frac{E_c h_0^2}{a^3 q_0}$ of CCCC and SSSS TD-FG SUS304/Si$_3$N$_4$ thick square plate $a/h_0=20$ and thin plate $a/h_0=50$.

| | | $k_z$ | $k_x$ / $k_y$ | 2 | | 4 | | 8 | |
|---|---|---|---|---|---|---|---|---|---|
| | | | | Present | Predict | Present | Predict | Present | Predict |
| Thick plates | CCCC | 2 | 2 | 0.3488 | 0.3640 | 0.3521 | 0.3587 | 0.3538 | 0.3461 |
| | | | 4 | 0.3521 | 0.3597 | 0.3540 | 0.3527 | 0.3550 | 0.3460 |
| | | | 8 | 0.3538 | 0.3568 | 0.3550 | 0.3529 | 0.3555 | 0.3500 |
| | | 6 | 2 | 0.3512 | 0.3599 | 0.3534 | 0.3570 | 0.3545 | 0.3525 |
| | | | 4 | 0.3534 | 0.3569 | 0.3546 | 0.3534 | 0.3553 | 0.3512 |
| | | | 8 | 0.3545 | 0.3555 | 0.3553 | 0.3537 | 0.3556 | 0.3513 |
| | SSSS | 2 | 2 | 0.9471 | 0.9459 | 0.9558 | 0.9666 | 0.9614 | 0.9633 |
| | | | 4 | 0.9558 | 0.9653 | 0.9602 | 0.9747 | 0.9634 | 0.9650 |
| | | | 8 | 0.9614 | 0.9717 | 0.9634 | 0.9744 | 0.9651 | 0.9620 |
| | | 6 | 2 | 0.9543 | 0.9747 | 0.9600 | 0.9836 | 0.9637 | 0.9791 |
| | | | 4 | 0.9600 | 0.9801 | 0.9629 | 0.9834 | 0.9650 | 0.9761 |
| | | | 8 | 0.9637 | 0.9756 | 0.9650 | 0.9751 | 0.9661 | 0.9687 |
| Thin plates | CCCC | 2 | 2 | 0.8379 | 0.8490 | 0.8458 | 0.8576 | 0.8500 | 0.8614 |
| | | | 4 | 0.8458 | 0.8554 | 0.8503 | 0.8603 | 0.8527 | 0.8643 |
| | | | 8 | 0.8500 | 0.8581 | 0.8527 | 0.8623 | 0.8541 | 0.8677 |
| | | 6 | 2 | 0.8436 | 0.8505 | 0.8489 | 0.8584 | 0.8516 | 0.8619 |
| | | | 4 | 0.8489 | 0.8586 | 0.8519 | 0.8623 | 0.8534 | 0.8634 |
| | | | 8 | 0.8516 | 0.8624 | 0.8534 | 0.8639 | 0.8543 | 0.8643 |
| | SSSS | 2 | 2 | 2.3385 | 2.3738 | 2.3601 | 2.3985 | 2.3739 | 2.4038 |
| | | | 4 | 2.3601 | 2.4023 | 2.3710 | 2.4135 | 2.3789 | 2.4124 |
| | | | 8 | 2.3739 | 2.4158 | 2.3789 | 2.4200 | 2.3830 | 2.4169 |
| | | 6 | 2 | 2.3562 | 2.3851 | 2.3704 | 2.4020 | 2.3795 | 2.4143 |
| | | | 4 | 2.3704 | 2.4028 | 2.3776 | 2.4104 | 2.3828 | 2.4165 |
| | | | 8 | 2.3795 | 2.4130 | 2.3828 | 2.4146 | 2.3855 | 2.4179 |

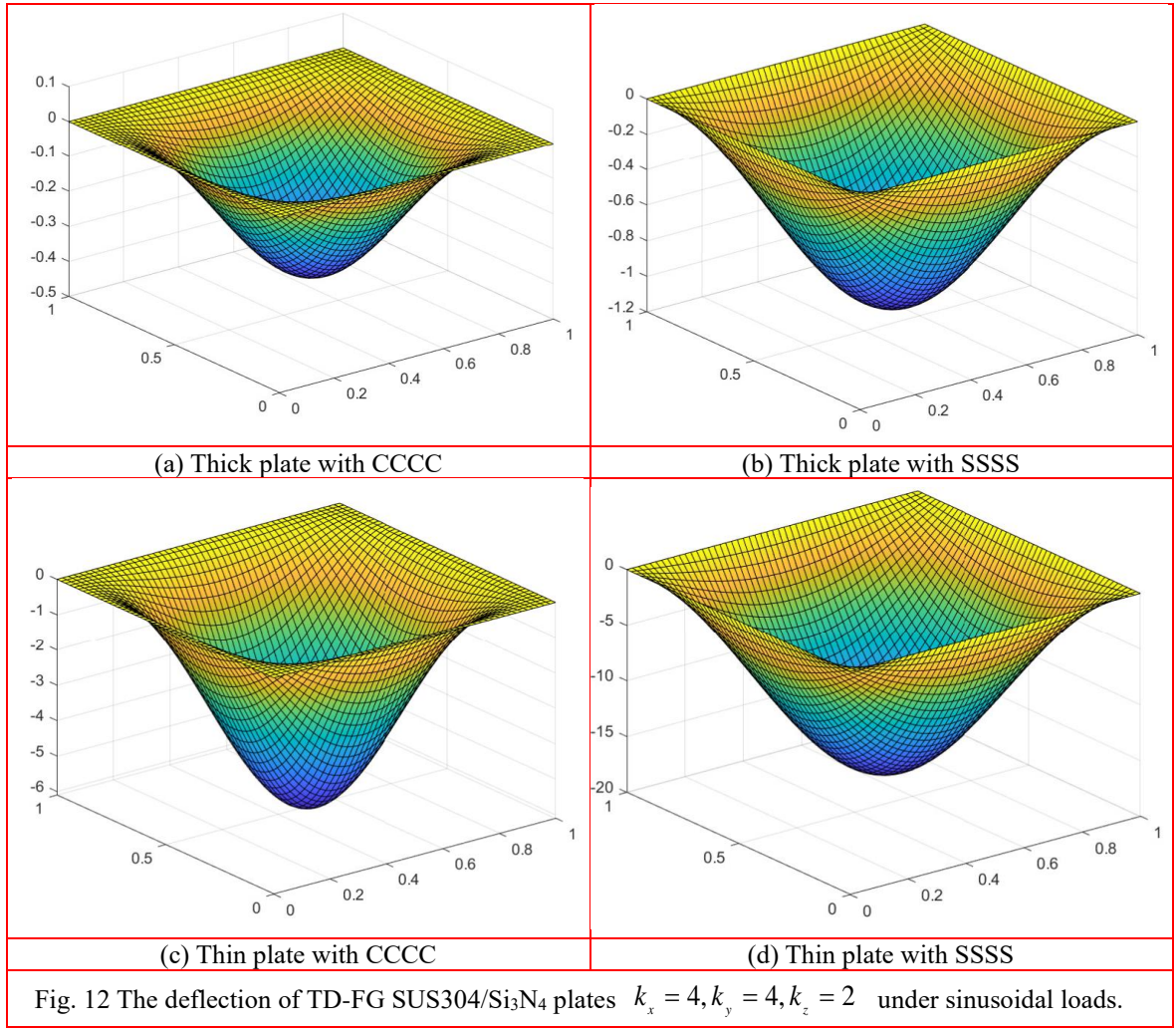

(a) Thick plate with CCCC  (b) Thick plate with SSSS
(c) Thin plate with CCCC  (d) Thin plate with SSSS

Fig. 12 The deflection of TD-FG SUS304/Si$_3$N$_4$ plates $k_x = 4, k_y = 4, k_z = 2$ under sinusoidal loads.

### 5.2 Free-vibration problems

In this part, the free vibration examples are investigated in terms of both analysis and prediction problems.

**Example 4:** The Al/Al$_2$O$_3$ UD-FG plate is investigated with both CFCF and CFFF boundary conditions. In this example, the non-dimensional frequency $\bar{\omega} = \omega h_0 \sqrt{\rho_m / E_m}$ is applied. The material distribution of FG properties is estimated by the rule of mixture rule. Fig. 13 shows the comparison with those of three-dimensional quadrature element method Wang et al. (2019). From the results, it can be seen that the presented frequencies are in good agreement with the previous results with very small error.

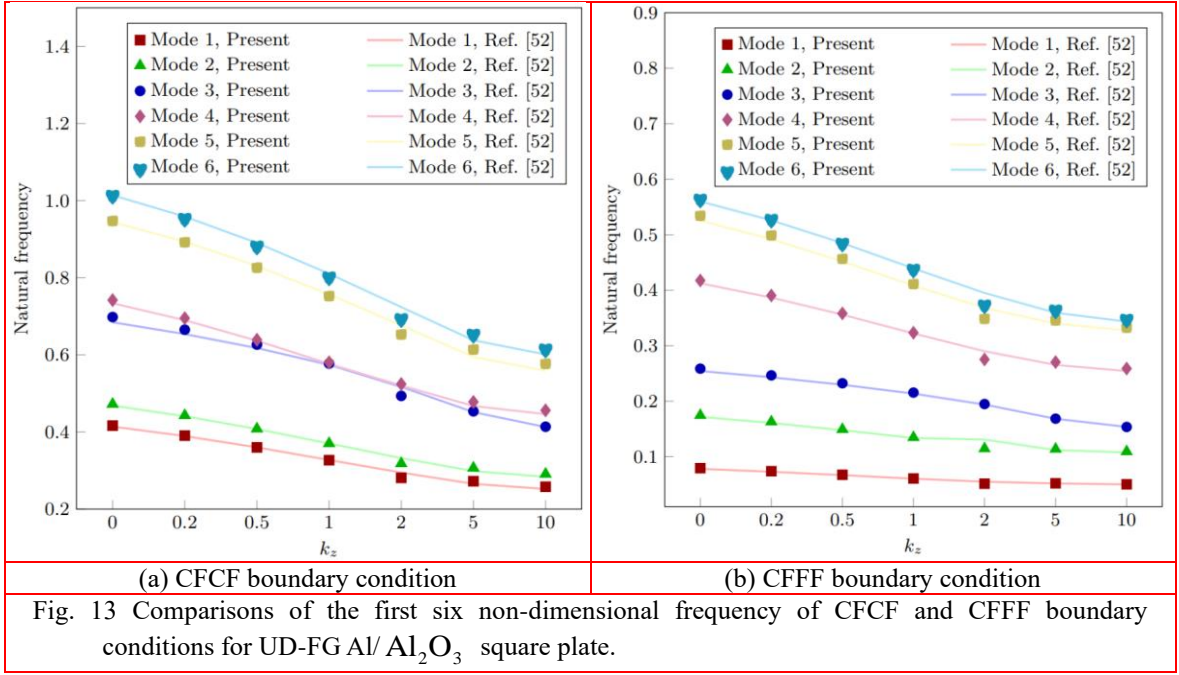

(a) CFCF boundary condition

(b) CFFF boundary condition

Fig. 13 Comparisons of the first six non-dimensional frequency of CFCF and CFFF boundary conditions for UD-FG Al/$Al_2O_3$ square plate.

**Example 5:** This example is currently dedicated to examining the free vibration responses of in-plane BD-FG plates ($k_z$=0) using a uniform thickness in the comparison reference Lieu et al. (2018). The thickness $h_0$ of the IBFG square plate is investigated in term of both thick (a/$h_0$=5) and thin (a/$h_0$=100) cases, wherein, a is the length of model. A SUS304/$Si_3N_4$ IBFG square plate with the non-dimensional frequency are defined Eq. (33b). Fig. 14 presents the first non-dimensional frequency of the previous problem with CCCC and SSSS boundary conditions in vary numbers of indexes $k_x$ and $k_y$. As observed, the present method can be concluded as effective and reliable for the BD-FG plates under free vibration. Furthermore, the accuracy of the current approach can ensure the generated dataset for the following training process in DNN.

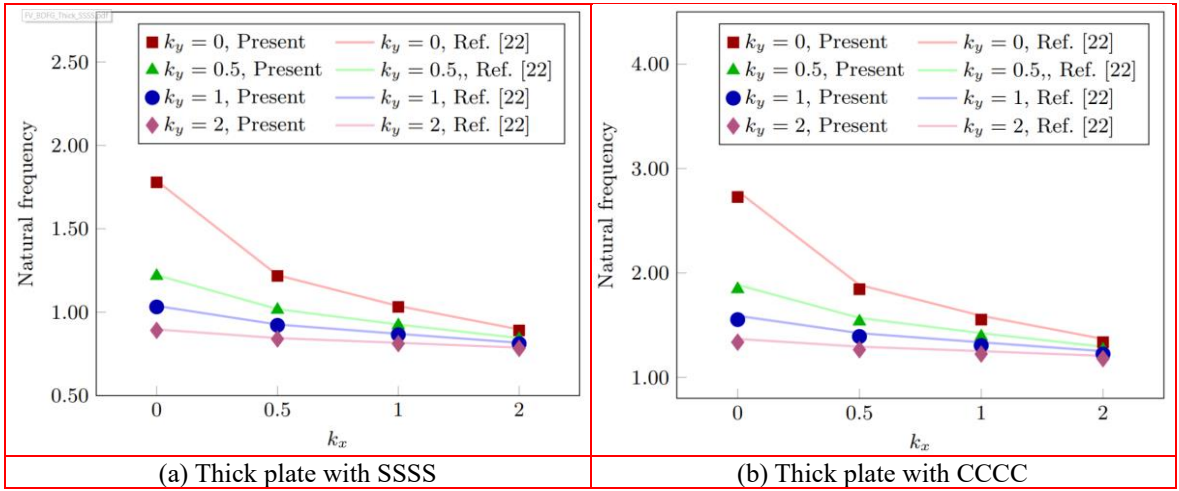

(a) Thick plate with SSSS

(b) Thick plate with CCCC

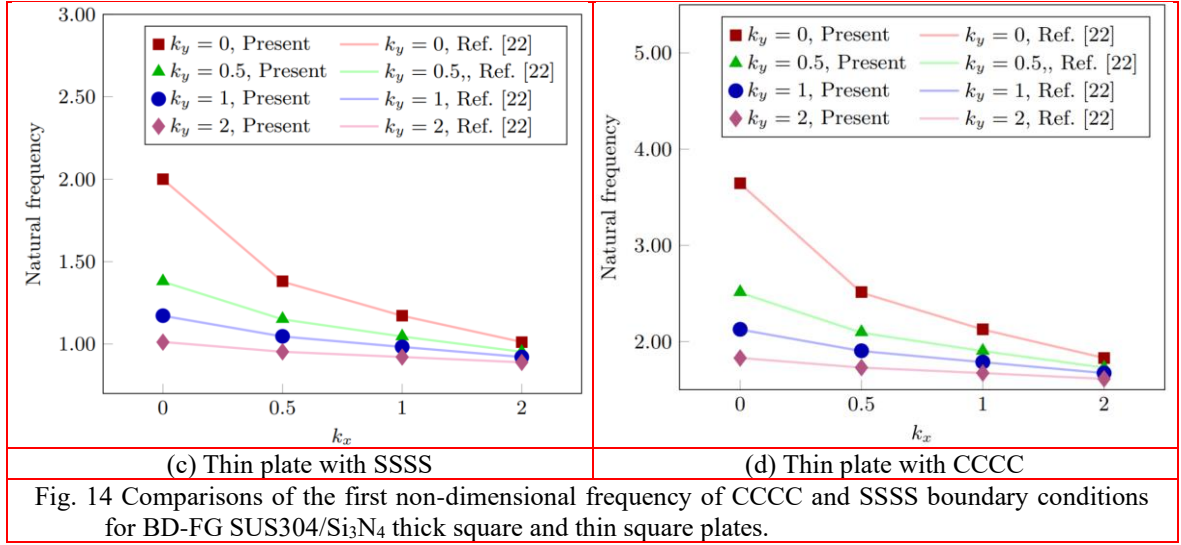

(c) Thin plate with SSSS     (d) Thin plate with CCCC

Fig. 14 Comparisons of the first non-dimensional frequency of CCCC and SSSS boundary conditions for BD-FG SUS304/$Si_3N_4$ thick square and thin square plates.

**Example 6:** In this example, the free vibration responses of both TD-FG thick and thin plates using three kind of thickness models: uniform (type 1), non-uniform linear (type 2), and non-uniform non-linear (type 3) are considered. A SUS304/$Si_3N_4$ TD-FG square plate with the non-dimensional frequency as defined Eq. (33b).

Tables 5, 6, 7 and 8 show the non-dimensional of first natural frequencies of TDFG plates with three types of thickness for CCCC and SSSS boundary conditions, respectively. As can be seen, the relative errors of all cases are smaller than 2.5% where the average relative errors of CCCC and SSSS boundary condition in free-vibration of TD-FG plate is 1.7551% and 1.0176%, respectively. On the other hand, the relative errors are 1.1611% (type 1), 1.6332% (type 2), and 1.3647% (type 3) in different type of thickness. In particular, from tables, it can be seen that when gradient indexes $k_x$, $k_y$, and $k_z$ increase, the contribution of the ceramic phase will decrease which leads to stiffness of the plate will decrease. As we know that a vibration frequency is directly proportional to the plate stiffness; therefore, the attained frequencies decrease when gradient indexes increase. Besides that, as can be seen from pair tables 5-7 and 6-8, the non-dimensional frequency increases when the length-to-thickness ratio increases. Figs. 15, 16, and 17 show the first three free vibration mode shapes of the SSSS boundary condition of TD-FG SUS304/$Si_3N_4$ square plate $a/h_0$=100 with $k_x$=5, $k_y$=5, $k_z$=2 in three types of thickness.

Table 5 Comparison studies of the first non-dimensional frequency $\bar{\omega} = \omega(a/\pi)^2 \sqrt{\rho_c h_0 / D_c}, D_c = E_c h_0^3 / 12(1-\nu_c^2)$ of CCCC boundary conditions for TD-FG SUS304/Si$_3$N$_4$ thick square plate a/h$_0$=10.

| | | | \multicolumn{8}{c|}{$k_x$} |
|---|---|---|---|---|---|---|---|---|---|---|
| | $k_z$ | | 1 | | 2 | | 5 | | 10 | |
| | | $k_y$ | Present | Predict | Present | Predict | Present | Predict | Present | Predict |
| Uniform | 1 | 1 | 1.4903 | 1.4732 | 1.4435 | 1.4373 | 1.4020 | 1.4019 | 1.3900 | 1.3862 |
| | | 2 | 1.4435 | 1.4340 | 1.4172 | 1.4139 | 1.3930 | 1.3945 | 1.3858 | 1.3833 |
| | | 5 | 1.4020 | 1.4051 | 1.3930 | 1.3976 | 1.3844 | 1.3867 | 1.3818 | 1.3798 |
| | | 10 | 1.3900 | 1.3908 | 1.3858 | 1.3871 | 1.3818 | 1.3830 | 1.3807 | 1.3760 |
| | 2 | 1 | 1.4579 | 1.4419 | 1.4256 | 1.4162 | 1.3964 | 1.3933 | 1.3877 | 1.3827 |
| | | 2 | 1.4256 | 1.4159 | 1.4070 | 1.4034 | 1.3897 | 1.3897 | 1.3845 | 1.3812 |
| | | 5 | 1.3964 | 1.3989 | 1.3897 | 1.3936 | 1.3834 | 1.3848 | 1.3815 | 1.3792 |
| | | 10 | 1.3877 | 1.3885 | 1.3845 | 1.3860 | 1.3815 | 1.3836 | 1.3806 | 1.3760 |
| Non-uniform linear | 1 | 1 | 1.9789 | 1.9643 | 1.9170 | 1.9031 | 1.8655 | 1.8489 | 1.8509 | 1.8326 |
| | | 2 | 1.9203 | 1.9007 | 1.8852 | 1.8674 | 1.8547 | 1.8376 | 1.8459 | 1.8271 |
| | | 5 | 1.8681 | 1.8542 | 1.8557 | 1.8441 | 1.8444 | 1.8272 | 1.8412 | 1.8229 |
| | | 10 | 1.8522 | 1.8418 | 1.8465 | 1.8371 | 1.8412 | 1.8291 | 1.8398 | 1.8259 |
| | 2 | 1 | 1.9374 | 1.9114 | 1.8948 | 1.8712 | 1.8586 | 1.8398 | 1.8480 | 1.8304 |
| | | 2 | 1.8971 | 1.8720 | 1.8723 | 1.8528 | 1.8507 | 1.8328 | 1.8443 | 1.8262 |
| | | 5 | 1.8604 | 1.8476 | 1.8513 | 1.8395 | 1.8431 | 1.8256 | 1.8407 | 1.8237 |
| | | 10 | 1.8490 | 1.8394 | 1.8447 | 1.8349 | 1.8407 | 1.8294 | 1.8396 | 1.8268 |
| Non-uniform non-linear | 1 | 1 | 1.7353 | 1.7262 | 1.6804 | 1.6817 | 1.6318 | 1.6332 | 1.6175 | 1.6154 |
| | | 2 | 1.6804 | 1.6746 | 1.6496 | 1.6477 | 1.6213 | 1.6191 | 1.6128 | 1.6097 |
| | | 5 | 1.6318 | 1.6321 | 1.6213 | 1.6203 | 1.6113 | 1.6067 | 1.6083 | 1.6047 |
| | | 10 | 1.6175 | 1.6165 | 1.6128 | 1.6110 | 1.6083 | 1.6075 | 1.6069 | 1.6043 |
| | 2 | 1 | 1.6972 | 1.6866 | 1.6594 | 1.6545 | 1.6253 | 1.6222 | 1.6149 | 1.6109 |
| | | 2 | 1.6594 | 1.6516 | 1.6377 | 1.6331 | 1.6175 | 1.6134 | 1.6113 | 1.6074 |
| | | 5 | 1.6253 | 1.6239 | 1.6175 | 1.6153 | 1.6101 | 1.6057 | 1.6078 | 1.6047 |
| | | 10 | 1.6149 | 1.6134 | 1.6113 | 1.6097 | 1.6078 | 1.6087 | 1.6068 | 1.6052 |

Table 6 Comparison studies of the first non-dimensional frequency $\bar{\omega} = \omega(a/\pi)^2 \sqrt{\rho_c h_0 / D_c}, D_c = E_c h_0^3 / 12(1-v_c^2)$ of CCCC boundary conditions for TD-FG SUS304/Si$_3$N$_4$ thick square plate a/h$_0$=10.

| $k_z$ | | $k_y$ | $k_x$ | | | | | | | |
|---|---|---|---|---|---|---|---|---|---|---|
| | | | 1 | | 2 | | 5 | | 10 | |
| | | | Present | Predict | Present | Predict | Present | Predict | Present | Predict |
| Uniform | 1 | 1 | 1.6527 | 1.6377 | 1.6009 | 1.5929 | 1.5552 | 1.5549 | 1.5423 | 1.5382 |
| | | 2 | 1.6009 | 1.5859 | 1.5720 | 1.5630 | 1.5455 | 1.5448 | 1.5378 | 1.5345 |
| | | 5 | 1.5552 | 1.5501 | 1.5455 | 1.5446 | 1.5363 | 1.5376 | 1.5336 | 1.5310 |
| | | 10 | 1.5423 | 1.5405 | 1.5378 | 1.5378 | 1.5336 | 1.5319 | 1.5324 | 1.5263 |
| | 2 | 1 | 1.6180 | 1.6053 | 1.5820 | 1.5750 | 1.5495 | 1.5482 | 1.5401 | 1.5358 |
| | | 2 | 1.5820 | 1.5685 | 1.5613 | 1.5539 | 1.5422 | 1.5405 | 1.5365 | 1.5327 |
| | | 5 | 1.5495 | 1.5462 | 1.5422 | 1.5415 | 1.5353 | 1.5346 | 1.5332 | 1.5300 |
| | | 10 | 1.5401 | 1.5388 | 1.5365 | 1.5358 | 1.5332 | 1.5303 | 1.5323 | 1.5264 |
| Non-uniform linear | 1 | 1 | 2.3973 | 2.4019 | 2.3229 | 2.3308 | 2.2631 | 2.2670 | 2.2475 | 2.2449 |
| | | 2 | 2.3284 | 2.3291 | 2.2866 | 2.2887 | 2.2515 | 2.2514 | 2.2420 | 2.2372 |
| | | 5 | 2.2671 | 2.2681 | 2.2530 | 2.2527 | 2.2404 | 2.2331 | 2.2369 | 2.2293 |
| | | 10 | 2.2494 | 2.2447 | 2.2429 | 2.2371 | 2.2370 | 2.2294 | 2.2354 | 2.2295 |
| | 2 | 1 | 2.3513 | 2.3453 | 2.2994 | 2.2976 | 2.2565 | 2.2540 | 2.2449 | 2.2376 |
| | | 2 | 2.3031 | 2.2961 | 2.2731 | 2.2699 | 2.2476 | 2.2417 | 2.2406 | 2.2314 |
| | | 5 | 2.2594 | 2.2559 | 2.2487 | 2.2440 | 2.2391 | 2.2286 | 2.2365 | 2.2269 |
| | | 10 | 2.2464 | 2.2420 | 2.2412 | 2.2360 | 2.2365 | 2.2306 | 2.2353 | 2.2305 |
| Non-uniform non-linear | 1 | 1 | 1.9719 | 1.9609 | 1.9096 | 1.9038 | 1.8547 | 1.8519 | 1.8391 | 1.8315 |
| | | 2 | 1.9096 | 1.8921 | 1.8750 | 1.8623 | 1.8432 | 1.8373 | 1.8340 | 1.8252 |
| | | 5 | 1.8547 | 1.8408 | 1.8432 | 1.8338 | 1.8323 | 1.8257 | 1.8290 | 1.8202 |
| | | 10 | 1.8391 | 1.8311 | 1.8340 | 1.8290 | 1.8290 | 1.8256 | 1.8276 | 1.8205 |
| | 2 | 1 | 1.9304 | 1.9098 | 1.8871 | 1.8743 | 1.8481 | 1.8435 | 1.8366 | 1.8278 |
| | | 2 | 1.8871 | 1.8646 | 1.8623 | 1.8489 | 1.8393 | 1.8336 | 1.8325 | 1.8230 |
| | | 5 | 1.8481 | 1.8367 | 1.8393 | 1.8325 | 1.8311 | 1.8258 | 1.8286 | 1.8200 |
| | | 10 | 1.8366 | 1.8335 | 1.8325 | 1.8317 | 1.8286 | 1.8276 | 1.8275 | 1.8211 |

Table 7 Comparison studies of the first non-dimensional frequency $\bar{\omega} = \omega(a/\pi)^2 \sqrt{\rho_c h_0 / D_c}, D_c = E_c h_0^3 / 12(1-v_c^2)$ of SSSS boundary conditions for TD-FG SUS304/Si$_3$N$_4$ thick square plate a/h$_0$=10.

| $k_z$ | | $k_y$ | $k_x$ | | | | | | | |
|---|---|---|---|---|---|---|---|---|---|---|
| | | | 1 | | 2 | | 5 | | 10 | |
| | | | Present | Predict | Present | Predict | Present | Predict | Present | Predict |
| Uniform | 1 | 1 | 0.8762 | 0.8743 | 0.8492 | 0.8529 | 0.8248 | 0.8339 | 0.8166 | 0.8232 |
| | | 2 | 0.8492 | 0.8500 | 0.8341 | 0.8411 | 0.8199 | 0.8322 | 0.8149 | 0.8240 |
| | | 5 | 0.8248 | 0.8420 | 0.8199 | 0.8384 | 0.8151 | 0.8320 | 0.8131 | 0.8235 |
| | | 10 | 0.8166 | 0.8389 | 0.8149 | 0.8358 | 0.8131 | 0.8292 | 0.8122 | 0.8207 |
| | 2 | 1 | 0.8538 | 0.8515 | 0.8388 | 0.8399 | 0.8213 | 0.8296 | 0.8154 | 0.8227 |
| | | 2 | 0.8388 | 0.8421 | 0.8280 | 0.8370 | 0.8177 | 0.8304 | 0.8140 | 0.8238 |
| | | 5 | 0.8213 | 0.8395 | 0.8178 | 0.8361 | 0.8142 | 0.8298 | 0.8127 | 0.8230 |
| | | 10 | 0.8154 | 0.8345 | 0.8140 | 0.8318 | 0.8127 | 0.8266 | 0.8120 | 0.8204 |
| Non-uniform linear | 1 | 1 | 1.2433 | 1.2414 | 1.2051 | 1.2007 | 1.1716 | 1.1638 | 1.1605 | 1.1478 |
| | | 2 | 1.2066 | 1.1951 | 1.1850 | 1.1730 | 1.1651 | 1.1555 | 1.1580 | 1.1463 |
| | | 5 | 1.1730 | 1.1584 | 1.1659 | 1.1532 | 1.1585 | 1.1488 | 1.1553 | 1.1457 |
| | | 10 | 1.1613 | 1.1483 | 1.1659 | 1.1476 | 1.1555 | 1.1488 | 1.1539 | 1.1449 |
| | 2 | 1 | 1.2119 | 1.1978 | 1.1872 | 1.1737 | 1.1651 | 1.1552 | 1.1576 | 1.1463 |
| | | 2 | 1.1882 | 1.1730 | 1.1739 | 1.1623 | 1.1607 | 1.1521 | 1.1560 | 1.1462 |
| | | 5 | 1.1660 | 1.1570 | 1.1612 | 1.1526 | 1.1563 | 1.1480 | 1.1542 | 1.1465 |
| | | 10 | 1.1581 | 1.1501 | 1.1563 | 1.1484 | 1.1543 | 1.1491 | 1.1532 | 1.1464 |
| Non-uniform non-linear | 1 | 1 | 1.0214 | 1.0438 | 0.9902 | 1.0129 | 0.9618 | 0.9821 | 0.9520 | 0.9662 |
| | | 2 | 0.9902 | 0.9993 | 0.9726 | 0.9843 | 0.9559 | 0.9726 | 0.9497 | 0.9637 |
| | | 5 | 0.9618 | 0.9740 | 0.9559 | 0.9707 | 0.9499 | 0.9688 | 0.9473 | 0.9637 |
| | | 10 | 0.9520 | 0.9667 | 0.9497 | 0.9656 | 0.9473 | 0.9646 | 0.9460 | 0.9639 |
| | 2 | 1 | 0.9948 | 0.9995 | 0.9746 | 0.9833 | 0.9559 | 0.9724 | 0.9493 | 0.9656 |
| | | 2 | 0.9746 | 0.9768 | 0.9631 | 0.9714 | 0.9520 | 0.9691 | 0.9478 | 0.9646 |
| | | 5 | 0.9559 | 0.9692 | 0.9520 | 0.9680 | 0.9480 | 0.9675 | 0.9462 | 0.9655 |
| | | 10 | 0.9493 | 0.9638 | 0.9478 | 0.9630 | 0.9462 | 0.9632 | 0.9454 | 0.9660 |

Table 8 Comparison studies of the first non-dimensional frequency $\bar{\omega} = \omega(a/\pi)^2 \sqrt{\rho_c h_0 / D_c}, D_c = E_c h_0^3 / 12(1 - v_c^2)$ of SSSS boundary conditions for TD-FG SUS304/Si$_3$N$_4$ thick square plate a/h$_0$=100.

| | $k_z$ | $k_y$ | $k_x$ | | | | | | | |
|---|---|---|---|---|---|---|---|---|---|---|
| | | | 1 | | 2 | | 5 | | 10 | |
| | | | Present | Predict | Present | Predict | Present | Predict | Present | Predict |
| Uniform | 1 | 1 | 0.9083 | 0.9294 | 0.8802 | 0.8989 | 0.8545 | 0.8642 | 0.8463 | 0.8519 |
| | | 2 | 0.8802 | 0.8914 | 0.8645 | 0.8737 | 0.8498 | 0.8540 | 0.8446 | 0.8474 |
| | | 5 | 0.8547 | 0.8641 | 0.8498 | 0.8573 | 0.8448 | 0.8492 | 0.8427 | 0.8456 |
| | | 10 | 0.8463 | 0.8575 | 0.8446 | 0.8542 | 0.8428 | 0.8491 | 0.8418 | 0.8459 |
| | 2 | 1 | 0.8888 | 0.9025 | 0.8692 | 0.8822 | 0.8511 | 0.8588 | 0.8450 | 0.8488 |
| | | 2 | 0.8692 | 0.8763 | 0.8581 | 0.8651 | 0.8475 | 0.8518 | 0.8437 | 0.8453 |
| | | 5 | 0.8511 | 0.8582 | 0.8475 | 0.8543 | 0.8439 | 0.8485 | 0.8423 | 0.8435 |
| | | 10 | 0.8450 | 0.8544 | 0.8437 | 0.8525 | 0.8423 | 0.8483 | 0.8416 | 0.8443 |
| Non-uniform linear | 1 | 1 | 1.3394 | 1.3438 | 1.2983 | 1.3096 | 1.2629 | 1.2672 | 1.2514 | 1.2540 |
| | | 2 | 1.3002 | 1.3007 | 1.2772 | 1.2811 | 1.2562 | 1.2574 | 1.2488 | 1.2502 |
| | | 5 | 1.2644 | 1.2674 | 1.2570 | 1.2596 | 1.2494 | 1.2492 | 1.2461 | 1.2464 |
| | | 10 | 1.2519 | 1.2580 | 1.2492 | 1.2550 | 1.2462 | 1.2513 | 1.2445 | 1.2423 |
| | 2 | 1 | 1.3059 | 1.2989 | 1.2793 | 1.2787 | 1.2559 | 1.2579 | 1.2482 | 1.2505 |
| | | 2 | 1.2806 | 1.2761 | 1.2654 | 1.2661 | 1.2515 | 1.2521 | 1.2465 | 1.2473 |
| | | 5 | 1.2569 | 1.2590 | 1.2520 | 1.2538 | 1.2469 | 1.2468 | 1.2447 | 1.2442 |
| | | 10 | 1.2486 | 1.2552 | 1.2468 | 1.2531 | 1.2448 | 1.2501 | 1.2437 | 1.2501 |
| Non-uniform non-linear | 1 | 1 | 1.0706 | 1.0850 | 1.0379 | 1.0478 | 1.0083 | 1.0143 | 0.9981 | 1.0070 |
| | | 2 | 1.0379 | 1.0385 | 1.0196 | 1.0206 | 1.0022 | 1.0073 | 0.9958 | 1.0045 |
| | | 5 | 1.0083 | 1.0108 | 1.0022 | 1.0059 | 0.9961 | 1.0032 | 0.9933 | 1.0000 |
| | | 10 | 0.9981 | 1.0019 | 0.9958 | 0.9998 | 0.9933 | 0.9974 | 0.9920 | 0.9933 |
| | 2 | 1 | 1.0428 | 1.0451 | 1.0217 | 1.0253 | 1.0021 | 1.0091 | 0.9953 | 1.0043 |
| | | 2 | 1.0217 | 1.0214 | 1.0096 | 1.0125 | 0.9981 | 1.0055 | 0.9938 | 1.0024 |
| | | 5 | 1.0021 | 1.0073 | 0.9981 | 1.0040 | 0.9940 | 1.0018 | 0.9921 | 0.9979 |
| | | 10 | 0.9953 | 0.9999 | 0.9938 | 0.9983 | 0.9921 | 0.9962 | 0.9912 | 0.9926 |

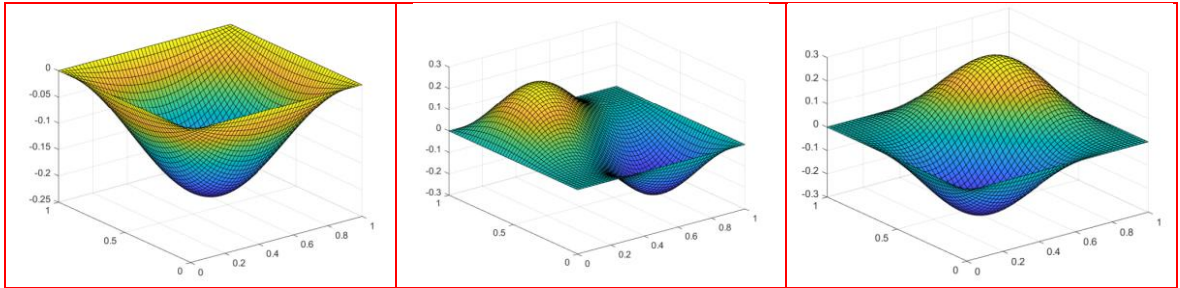

Fig. 15 The first three free vibration mode shapes of the SSSS TD-FG SUS304/$Si_3N_4$ thin square uniform (type 1) plate $a/h_0=100$ with $k_x=k_y=5$, $k_z=2$.

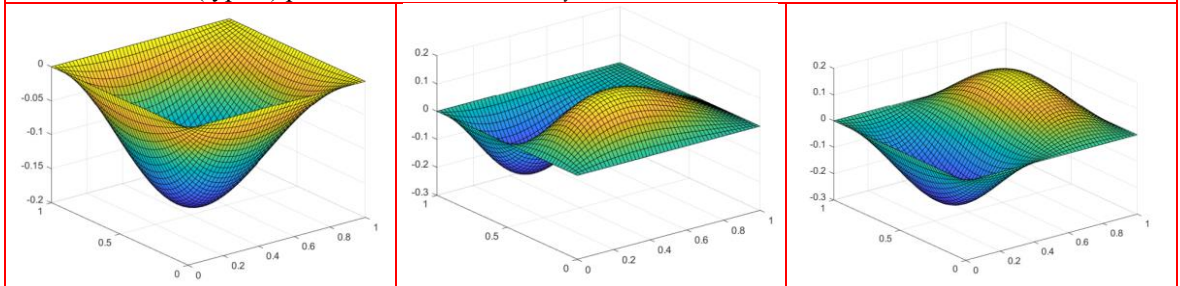

Fig. 16 The first three free vibration mode shapes of the SSSS TD-FG SUS304/$Si_3N_4$ thin square non-uniform linear (type 2) plate $a/h_0=100$ with $k_x=k_y=5$, $k_z=2$.

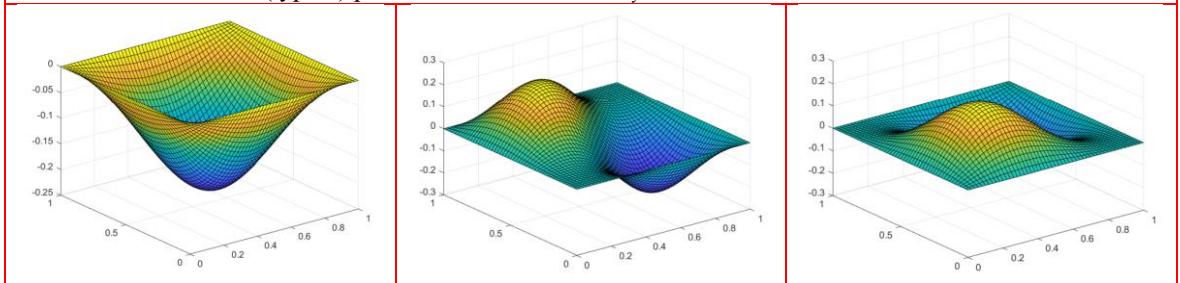

Fig. 17 The first three free vibration mode shapes of the SSSS TD-FG SUS304/$Si_3N_4$ thin square non-uniform non-linear (type 3) plate $a/h_0=100$ with $k_x=k_y=5$, $k_z=2$.

*5.3 Buckling problems*

The buckling investigations of the present model in terms of both analytical and predictive problems are presented in this part.

**Example 7:** The buckling analysis of UD-, BD-, and TD-FG square plates are compared to reference Do et al. (2020). The FG material SUS304/$Si_3N_4$ is utilized for these problems. Material properties of the FG materials are estimated by Mori-Tanaka scheme. The non-dimensional of uni-axial ($N^0_{xx} = 1$) and bi-axial ($N^0_{xx} = N^0_{yy} = 1$) critical buckling loads are defined in Eq. (33c).

In term of UD-FG material, Table 9 shows the comparisons of present study and DNN prediction with reference results Do et al. (2020) in uni-axial and bi-axial buckling problems with SSSS boundary condition. With CCCC boundary condition, the present results will be shown with the predicting results of DNN model. It can be seen that the present study is close to reference results and the DNN model predicts quite exactly. The relative errors of present results and DNN predictions are always smaller than 2% for both CCCC and SSSS and both uni-axial and bi-axial buckling problems. These results claim that the finite element model implemented for UD-FG plate is verified and the DNN model which is learned by using data created from the finite element analysis predicts with high accuracy.

Table 9 Comparisons and present study of the first non-dimensional critical buckling loads $\bar{P}_{cr} = \dfrac{P_{cr} a^2}{\pi^2 D_c}$ for UD-FG SUS304/$Si_3N_4$ square plate.

| | | $a/h_0$ | | $k_z$ | | | |
|---|---|---|---|---|---|---|---|
| | | | | 0 | 1 | 2 | 5 |
| Uni-axial buckling | SSSS | 10 | Ref. Do et al. (2020) | 3.8026 | 2.9522 | 2.7995 | 2.6511 |
| | | | Present | 3.8020 | 2.9917 | 2.8400 | 2.6722 |
| | | | Predict | 3.7939 | 3.0698 | 2.8977 | 2.6757 |
| | | 100 | Ref. Do et al. (2020) | 3.9979 | 3.1121 | 2.9627 | 2.8168 |
| | | | Present | 4.0001 | 3.1570 | 3.0052 | 2.8354 |
| | | | Predict | 3.9999 | 3.1782 | 3.0217 | 2.8712 |
| | CCCC | 10 | Present | 8.4122 | 6.4972 | 6.1262 | 5.7586 |
| | | | Predict | 8.4135 | 6.4381 | 6.1750 | 5.7676 |
| | | 100 | Present | 10.0786 | 7.8449 | 7.4636 | 7.0869 |
| | | | Predict | 10.0679 | 7.9904 | 7.5220 | 7.1285 |
| Bi-axial buckling | SSSS | 10 | Ref. Do et al. (2020) | 1.9013 | 1.4761 | 1.3998 | 1.3255 |
| | | | Present | 1.9010 | 1.4958 | 1.4200 | 1.3361 |
| | | | Predict | 1.9126 | 1.5174 | 1.4391 | 1.3385 |
| | | 100 | Ref. Do et al. (2020) | 1.9990 | 1.5560 | 1.4814 | 1.4084 |
| | | | Present | 2.0001 | 1.5785 | 1.5026 | 1.4177 |
| | | | Predict | 2.0000 | 1.5633 | 1.5233 | 1.4149 |
| | CCCC | 10 | Present | 4.5997 | 3.5586 | 3.3621 | 3.1675 |
| | | | Predict | 4.6113 | 3.7506 | 3.3186 | 3.2094 |
| | | 100 | Present | 5.3059 | 4.1301 | 3.9294 | 3.7312 |
| | | | Predict | 5.3257 | 4.2168 | 3.8595 | 3.7588 |

Fig. 18 compares the present studies of non-dimensional critical buckling load $\bar{P}_{cr}$ with those of reference results Do et al. (2020) with SSSS boundary condition for both thick and thin plates. In term of CCCC boundary condition, the prediction of non-dimensional critical buckling load $\bar{P}_{cr}$ for CCCC boundary condition is presented in Tab. 10 to compare with present results in case of BD-FG plate.

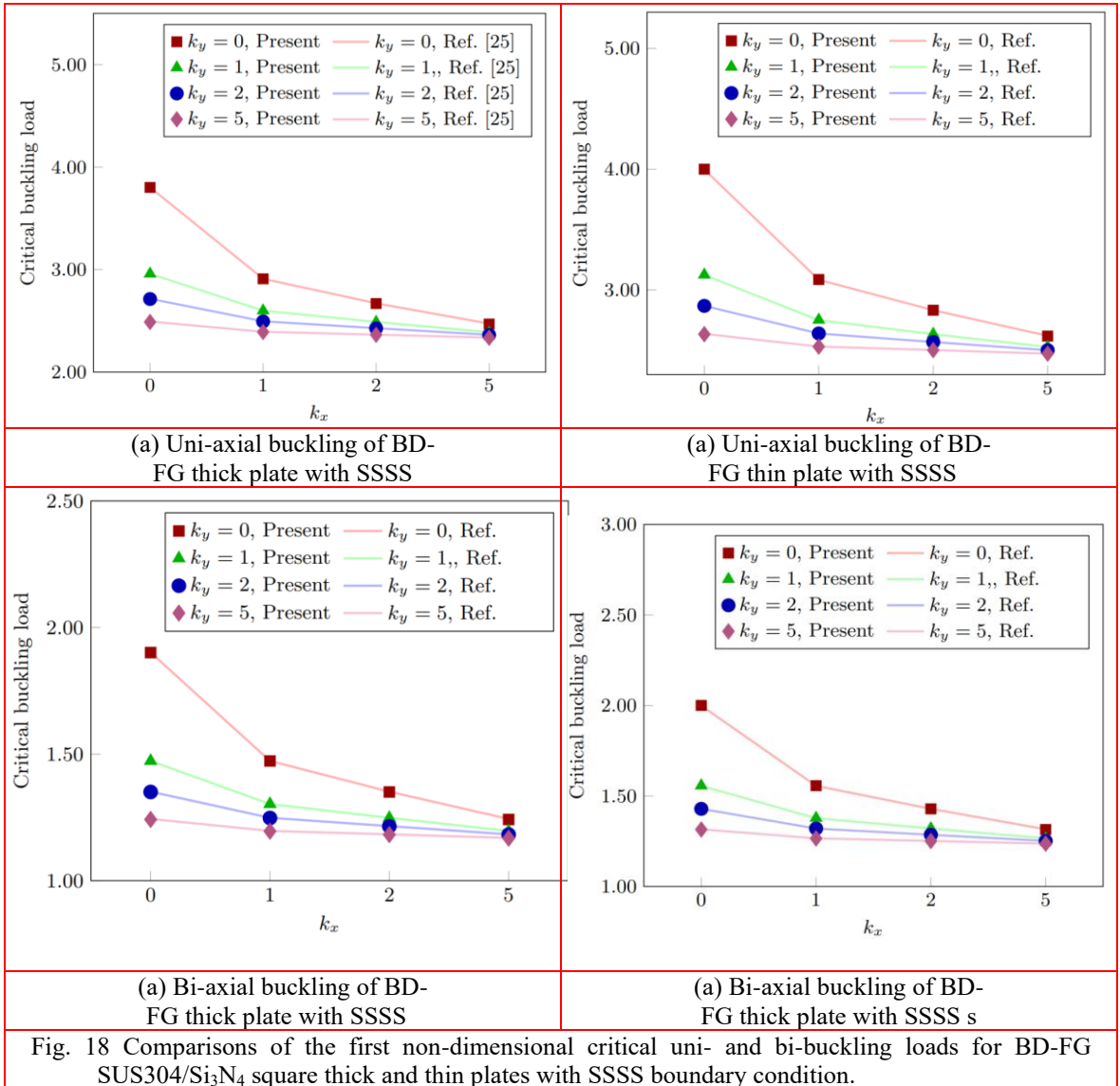

Fig. 18 Comparisons of the first non-dimensional critical uni- and bi-buckling loads for BD-FG SUS304/Si$_3$N$_4$ square thick and thin plates with SSSS boundary condition.

Table 10 Present study of the first non-dimensional critical buckling loads $\bar{P}_{cr} = \dfrac{P_{cr}a^2}{\pi^2 D_c}$ for BD-FG SUS304/Si$_3$N$_4$ square plate with CCCC boundary condition.

| | Results | $k_y$ | a/h$_0$=10 | | | a/h$_0$=10 | | |
|---|---|---|---|---|---|---|---|---|
| | | | $k_x$ | | | $k_x$ | | |
| | | | 1 | 5 | 10 | 1 | 5 | 10 |
| Uni-axis | Present | 1 | 5.6584 | 5.3713 | 5.1274 | 6.9119 | 6.5737 | 6.2798 |
| | Predict | | 5.5605 | 5.1451 | 5.0639 | 6.8754 | 6.3063 | 6.1832 |
| | Present | 2 | 5.4215 | 5.2428 | 5.0841 | 6.6300 | 6.4163 | 6.2243 |
| | Predict | | 5.4137 | 5.0871 | 5.0449 | 6.6626 | 6.2301 | 6.1544 |
| | Present | 5 | 5.1971 | 5.1154 | 5.0401 | 6.3630 | 6.2619 | 6.1688 |
| | Predict | | 5.2076 | 5.0450 | 5.0302 | 6.4030 | 6.1764 | 6.1363 |
| Bi-axis | Present | 1 | 3.1155 | 2.9729 | 2.8430 | 3.6455 | 3.4832 | 3.3347 |
| | Predict | | 3.0695 | 2.8831 | 2.8203 | 3.5483 | 3.3799 | 3.2932 |
| | Present | 2 | 2.9729 | 2.8872 | 2.8064 | 3.4832 | 3.3843 | 3.2913 |
| | Predict | | 2.9970 | 2.8367 | 2.8021 | 3.4806 | 3.3136 | 3.2672 |
| | Present | 5 | 2.8430 | 2.8064 | 2.7713 | 3.3347 | 3.2913 | 3.2499 |
| | Predict | | 2.8888 | 2.7988 | 2.7851 | 3.3638 | 3.2612 | 3.2452 |

Similarly, in TD-FG material, the behaviors of present study are compared with reference results in case of SSSS boundary condition while those of CCCC are predicted and verified with present studies in Tab. 11. Although the DNN predictions are always far from reference results than those of present results, the predictions are acceptable due to quite small relative errors and very quick time consumption of DNN model. Finally, Figs. 19-20 shows the three first buckling mode of TD-FG plates with uniform thickness for CCCC boundary condition.

Table 11 Present study of the first non-dimensional critical buckling loads $\bar{P}_{cr} = \dfrac{P_{cr}a^2}{\pi^2 D_c}$ for BD-FG SUS304/Si$_3$N$_4$ square plate with CCCC boundary condition.

| | | a/h$_0$ | Results | k$_z$ | | | |
|---|---|---|---|---|---|---|---|
| | | | | 0 | 1 | 2 | 5 |
| Uni-axial | SSSS | 10 | Ref. Do et al. (2020) | 2.6003 | 2.4470 | 2.4155 | 2.3819 |
| | | | Present | 2.5997 | 2.4503 | 2.4192 | 2.3835 |
| | | | Predict | 2.6325 | 2.4999 | 2.4402 | 2.3791 |
| | | 100 | Ref. Do et al. (2020) | 2.7509 | 2.5877 | 2.5564 | 2.5228 |
| | | | Present | 2.7524 | 2.5934 | 2.5618 | 2.5253 |
| | | | Predict | 2.8029 | 2.6506 | 2.5808 | 2.5423 |
| | CCCC | 10 | Present | 5.6584 | 5.3354 | 5.2615 | 5.1814 |
| | | | Predict | 5.5605 | 5.3801 | 5.2531 | 5.1680 |
| | | 100 | Present | 6.9119 | 6.5170 | 6.4390 | 6.3532 |
| | | | Predict | 6.8754 | 6.6038 | 6.4659 | 6.3806 |
| Bi-axial | SSSS | 10 | Ref. Do et al. (2020) | 1.3027 | 1.2241 | 1.2081 | 1.1911 |
| | | | Present | 1.3024 | 1.2258 | 1.2100 | 1.1920 |
| | | | Predict | 1.3279 | 1.2750 | 1.2418 | 1.2177 |
| | | 100 | Ref. Do et al. (2020) | 1.3772 | 1.2943 | 1.2785 | 1.2615 |
| | | | Present | 1.3780 | 1.2971 | 1.2812 | 1.2628 |
| | | | Predict | 1.3910 | 1.3374 | 1.3031 | 1.2735 |
| | CCCC | 10 | Present | 3.1155 | 2.9318 | 2.8916 | 2.8483 |
| | | | Predict | 3.0695 | 2.9872 | 2.9247 | 2.8761 |
| | | 100 | Present | 3.6455 | 3.4326 | 3.3910 | 3.3454 |
| | | | Predict | 3.5483 | 3.4661 | 3.4023 | 3.3529 |

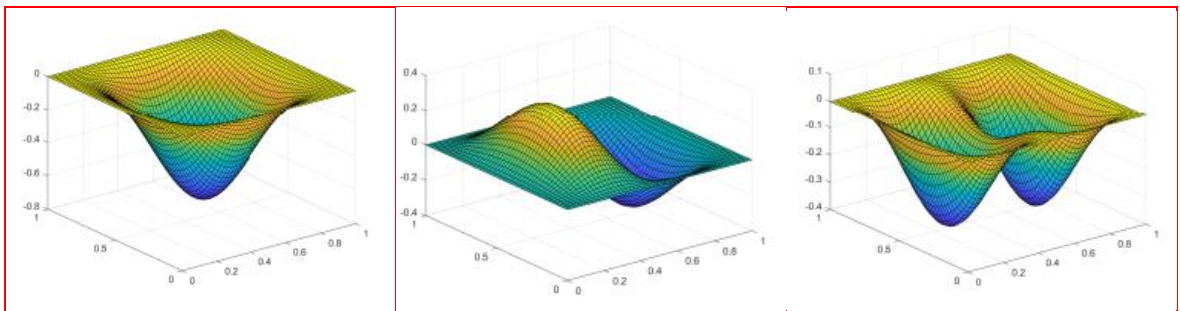

Fig. 19 The first three uni-axial buckling mode shapes of the CCCC TD-FG SUS304/Si$_3$N$_4$ thin uniform square plate a/h0 = 100 with k$_x$ = k$_y$ = 1, k$_z$ = 5.

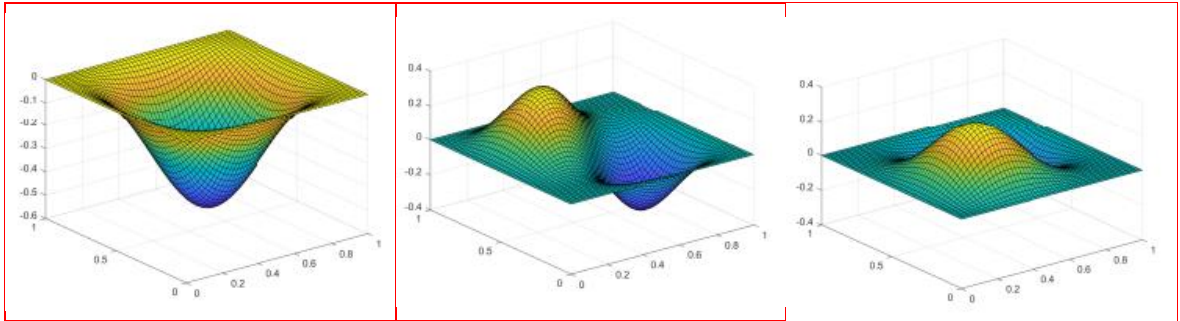

Fig. 20 The first three bi-axial buckling mode shapes of the CCCC TD-FG SUS304/Si$_3$N$_4$ thin uniform square uniform (type 1) plate a/h0 = 100 with $k_x = k_y = 1$, $k_z = 5$.

*5.4 TD-FG plates embedded into elastic foundation*

In this part, the influences of elastic foundation in the TD-FG plate are investigated in terms of both analytical and predictive approaches.

**Example 8:** A free-vibration problem of a UD-FG Al/Al$_2$O$_3$ with elastic Winkler foundation is considered. In this example, the non-dimensional natural frequency and Winkler parameter are defined as follows:

$$\bar{\omega} = \omega h_0 \sqrt{\rho_m / E_m} \text{ and } \bar{k}_W = k_W a^4 / D_{11} \text{ where } D_{11} = \int_{-h_0/2}^{h_0/2} z^2 Q_{11} dz$$

Fig. 21 shows the comparison of present and reference results TSDT (Baferani *et al.* 2011) and HSDT (Li *et al.* 2021). As expected, a closing agreement between gained outcomes and reference solutions is found. This once again demonstrates the reliability of present approach.

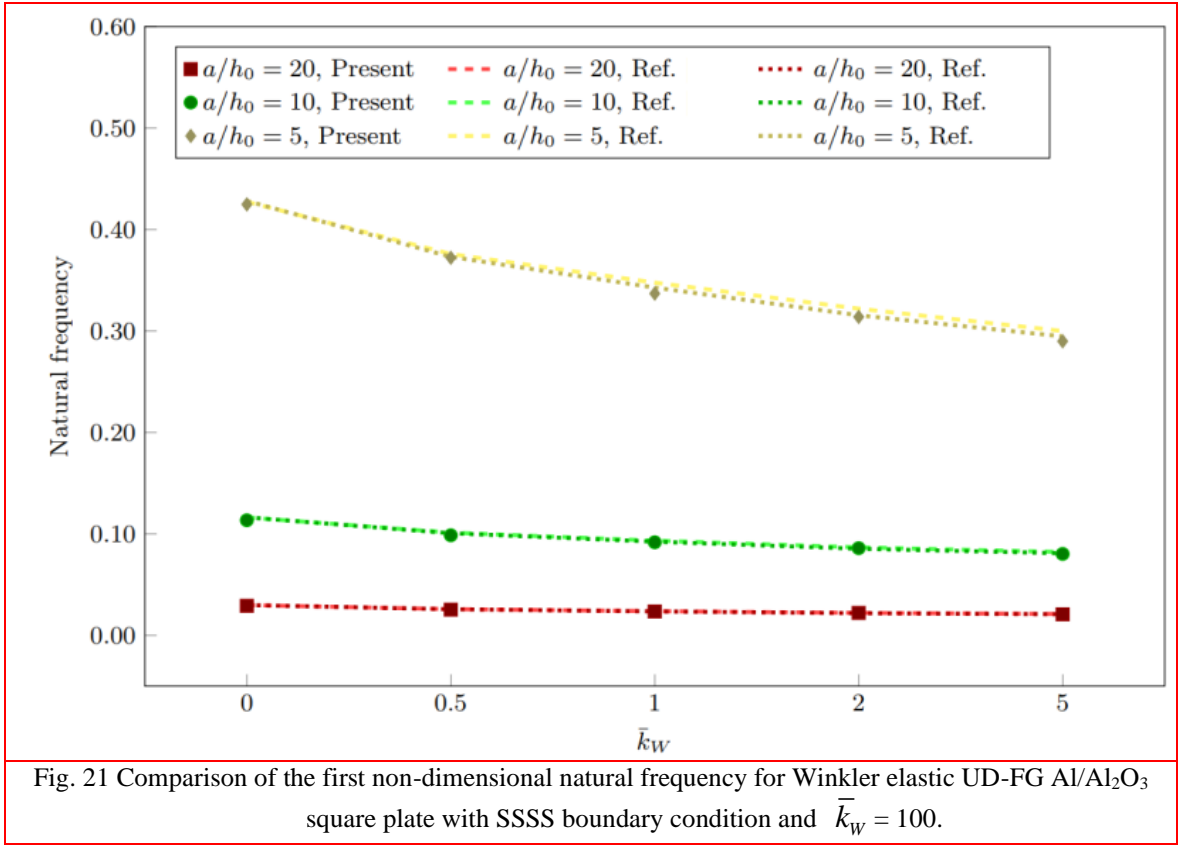

Fig. 21 Comparison of the first non-dimensional natural frequency for Winkler elastic UD-FG Al/Al$_2$O$_3$ square plate with SSSS boundary condition and $\bar{k}_W = 100$.

**Example 9:** A TD-FG ($k_x = k_z = 1$) SUS304/Si$_3$N$_4$ square plate with SSSS boundary condition is applied for an elastic Winkler elastic $\bar{k}_W = 3^4$ for bending, uni-axial and bi-axial buckling problems. The non-dimensional values are defined

$$\bar{w}\left(\frac{a}{2},\frac{a}{2},0\right) = \frac{E_c h_0^2}{a^3 q_0} w\left(\frac{a}{2},\frac{a}{2},0\right) \text{ and } \bar{P}_{cr} = \frac{P_{cr} a^2}{\pi^2 D_c}$$

for central deflection of bending problem and critical buckling load of both uni- and bi-axial buckling problem. Table 12 shows the comparison of present study and DNN prediction for both SSSS and CCCC boundary conditions. Small gaps of present study and prediction illustrate the accuracy of DNN model for TD-FG plates resting on elastic Winkler foundation.

Table 12 Present study of the first non-dimensional central deflection and critical buckling load for Winkler elastic TD-FG ($k_x = k_z = 1$) SUS304/Si$_3$N$_4$ plate with SSSS boundary condition and $k_W = 3^4$.

| | $a/h_0$ | | Bending | | Uni-axial buckling | | Bi-axial buckling | |
|---|---|---|---|---|---|---|---|---|
| | | $k_z$ | 1 | 5 | 1 | 5 | 1 | 5 |
| SSSS | 10 | Present | 0.4733 | 0.4871 | 2.4503 | 2.3835 | 1.2258 | 1.1920 |
| | | Predict | 0.4630 | 0.5053 | 2.5027 | 2.3881 | 1.2714 | 1.2182 |
| | 20 | Present | 0.9073 | 0.9327 | 2.5572 | 2.4894 | 1.2791 | 1.2449 |
| | | Predict | 0.8689 | 0.9344 | 2.5520 | 2.4703 | 1.2951 | 1.2484 |
| | 50 | Present | 2.2406 | 2.3029 | 2.5888 | 2.5207 | 1.2948 | 1.2605 |
| | | Predict | 2.2291 | 2.3309 | 2.6187 | 2.5162 | 1.3371 | 1.2753 |
| | 100 | Present | 4.4734 | 4.5974 | 2.5934 | 2.5253 | 1.2971 | 1.2628 |
| | | Predict | 4.3694 | 4.6058 | 2.6506 | 2.5439 | 1.3325 | 1.2714 |
| CCCC | 10 | Present | 0.1891 | 0.1948 | 5.3354 | 5.1814 | 2.9318 | 2.8483 |
| | | Predict | 0.1875 | 0.1924 | 5.3793 | 5.1663 | 2.9823 | 2.8707 |
| | 20 | Present | 0.3341 | 0.3434 | 6.1846 | 6.0226 | 3.2926 | 3.2061 |
| | | Predict | 0.3355 | 0.3553 | 6.2407 | 6.0088 | 3.3181 | 3.2187 |
| | 50 | Present | 0.8029 | 0.8245 | 6.4732 | 6.3096 | 3.4142 | 3.3271 |
| | | Predict | 0.8167 | 0.8305 | 6.5743 | 6.3188 | 3.4667 | 3.3358 |
| | 100 | Present | 1.5965 | 1.6392 | 6.5170 | 6.3532 | 3.4326 | 3.3454 |
| | | Predict | 1.5764 | 1.6275 | 6.5820 | 6.3713 | 3.4643 | 3.3506 |

## 6. Conclusion and remarks

A multi-directional FG square plate model with variable thickness resting on an elastic Winkler foundation regarding bending, free vibration, and buckling problems was described in mathematically detail. The Mori-Tanaka micro-mechanical technique is applied to describe the material properties that vary continuously through the plates' one, two, and three directions. The numerical performance is conducted to verify the effectiveness and reliability for analyzing the behavior of the present FG plate model in terms of all static, dynamic, and stability problems. Then, the DNN model using batch normalization and Adam optimizer is created to predict the non-dimensional values such as central deflection, natural frequency, and critical buckling load based on the dataset collected from the analytical solution. Several conclusions can be drawn via the presented formulations and examined examples as follows:

- The finite element analysis with MITC4 yields excellent results in comparison with those previous in the open literature for analyzing bending, free vibration and buckling behavior of the TD-FG plates.
- There are also several new results are presented in TD-FG plates, especially new results for TD-FG plates embedded on elastic Winkler foundation with various types of thickness (type 1, 2 and 3).

- The training dataset is collected based on finite element model. This dataset is essential for building a DNN model and the DNN helps to find the optimal mapping rule by learning the relation between input data and output one.
- The non-dimensional central deflection (bending problem), natural frequency (free-vibration problem), and critical buckling load (uni-axial and bi-axial buckling problems) can be directly predicted with high accuracy without solving system of linear equations problems or eigenvalue problems. The prediction is compared to present study and reference to claim the high accuracy of DNN model.

The present method can be extended to complex engineering problems such as shells or cracked (discontinuous) plates. Furthermore, topology optimization also awaits further attention.

**Acknowledgments**